\documentclass{article}

\usepackage{iclr2027_conference,times}
\usepackage[T1]{fontenc}

\usepackage{amsmath,amsfonts,bm}

\def\eqref#1{equation~\ref{#1}}

\def\1{\bm{1}}

\DeclareMathAlphabet{\mathsfit}{\encodingdefault}{\sfdefault}{m}{sl}
\SetMathAlphabet{\mathsfit}{bold}{\encodingdefault}{\sfdefault}{bx}{n}

\usepackage{amsmath}
\usepackage{amssymb}
\usepackage{booktabs}
\usepackage{graphicx}
\usepackage{float}
\usepackage{algorithm}
\usepackage{algpseudocode}
\usepackage{tabularx}
\usepackage{microtype}
\usepackage{placeins}
\usepackage{xcolor}
\usepackage{colortbl}
\usepackage{xspace}
\usepackage{hyperref}
\usepackage{xurl}
\hypersetup{
  hidelinks,
  pdftitle={ALDER: Discovering the Laws of a World by Acting in It},
  pdfauthor={Teng Cao, Yu Deng, Quentin Delfosse, Kristian Kersting},
  pdfsubject={Preprint}
}

\title{ALDER: Discovering the Laws of a World\\by Acting in It}

\author{%
  Teng Cao\textsuperscript{1,3,5,}\thanks{These authors contributed equally.}%
  \textsuperscript{,}\thanks{Corresponding authors:
    \href{mailto:yu.deng@tu-darmstadt.de}{\texttt{yu.deng@tu-darmstadt.de}},
    \href{mailto:teng.cao@tu-darmstadt.de}{\texttt{teng.cao@tu-darmstadt.de}}.}%
  \quad
  Yu Deng\textsuperscript{1,3,5,*,$\dagger$}\quad
  Quentin Delfosse\textsuperscript{1,2,3}\quad
  Kristian Kersting\textsuperscript{1,3,4,5}%
}

\iclrfinalcopy
\renewcommand{\headrulewidth}{0pt}
\makeatletter
\renewcommand{\@maketitle}{%
  \def\@makefnmark{\hbox{\@textsuperscript{\normalfont\@thefnmark}}}%
  \begin{center}
    {\LARGE\bfseries \@title\par}
    \vspace{1em}
    {\normalsize \@author\par}
    \vspace{0.7em}
    {\small
      \textsuperscript{1}Technische Universit\"at Darmstadt\quad
      \textsuperscript{2}Intrinsic\quad
      \textsuperscript{3}CS Department, TU Darmstadt\\
      \textsuperscript{4}German Research Center for AI\quad
      \textsuperscript{5}The Hessian Center for AI\par
    }
  \end{center}
  \vspace{0.5em}
}
\makeatother

\newcommand{\method}{\textsc{Alder}\xspace}
\newcommand{\lead}[1]{\textbf{#1}}
\newcommand{\dataset}{\mathcal{D}}
\newcommand{\hypotheses}{\mathcal{H}}
\newcommand{\queries}{\mathcal{Q}}

\definecolor{alderrow}{RGB}{232,232,255}

\begin{document}

\maketitle

\begin{abstract}
Reliable world models should not only predict future states but express how
actions change the world in an explicit, transparent and testable form, such as equations.
Yet methods that rely on a fixed set of trajectories cannot distinguish equally good competing hypotheses, while searches over a fixed set of predefined candidates cannot discover equations outside the initial hypothesis space.
We introduce \method (Action-guided Law Discovery, Evaluation, and Revision),
a method that actively proposes novel experiments to test and revise
models.
Specifically, \method proposes parametric equations; a numerical optimizer fits
their coefficients; an independent verifier tests these candidates on held-out
data. To distinguish between competing valid hypotheses, a cost- and safety-aware selector queries interventions, in the form of novel experiments.
The resulting counterexamples update the
evidence ledger and guide the next structural revision, while incompatible laws
are discarded.
Across an in-house benchmark, ODE equation discovery tasks, and robotic
experiments, \method discovers laws beyond its initial formula set, repairs
failed model proposals, distinguishes fixed candidate models with fewer
interactions, and improves out-of-distribution prediction.
Furthermore, given a current state and a target, \method selects control actions
by solving the inverse problem defined by its validated world model.
Together, these results show that explicit equation-based world models can be tested and revised through interaction, then naturally used to guide goal-directed control.
\end{abstract}

\section{Introduction}

\lead{Autonomous agents need world models that can predict beyond their initial learning support.}
A world model must not only fit observed transitions but anticipate how untried
actions affect the environment. Explicit behavioral models such as ESBM
represent action consequences as inspectable and reusable structured knowledge
\citep{shindo2026esbm}. For physical and robotic tasks, models must also predict
action-induced changes in continuous states to support planning and control
\citep{hafner2020dreamer,hafner2019planet}. Explicit equations offer an
inspectable representation of dynamics \citep{brunton2016sindy,cranmer2023pysr}:
they describe how states and actions determine predicted changes, allowing
action outcomes to be simulated and predictions to be tested through new
interactions. Learning reliable world models represented by equations from
limited interaction is therefore an important problem connecting world
modeling with model-based control.

\lead{Accurate predictions on the support set of trajectories do not guarantee an exact world model.} Trajectories collected in limited scenarios typically cover only a small set of state--action combinations, on which structurally different equations may make nearly identical predictions.
When object properties, action ranges, or contact conditions change, however,
their predictions may diverge substantially. Fitting error on existing
trajectories alone therefore cannot determine which model will predict
reliably under new conditions; additional interactions are needed to test and
distinguish these models by comparing their predictions with observed outcomes.

\lead{World model learning must go beyond fixed data and beyond predefined equations.}
Sparse-library methods construct equations from predefined terms but cannot
introduce variables or operations absent from the library
\citep{brunton2016sindy,champion2019coordinates}. Symbolic regression can search
for new expressions
\citep{udrescu2020aifeynman,cranmer2023pysr,shojaee2025llmsr,xia2026srscientist},
but when existing data cannot distinguish candidate models, expanding the
search alone cannot determine which model will predict correctly under new
conditions. Active system identification can choose inputs to collect
additional information, but formulations with prespecified model structures
do not add missing equation terms \citep{wagenmaker2020active}. Addressing both
problems therefore calls for experiments that distinguish candidate models
and the ability to revise equation structures when fitted models fail to
explain the observations.

\begin{figure}[!t]
  \centering
  \includegraphics[width=\textwidth]{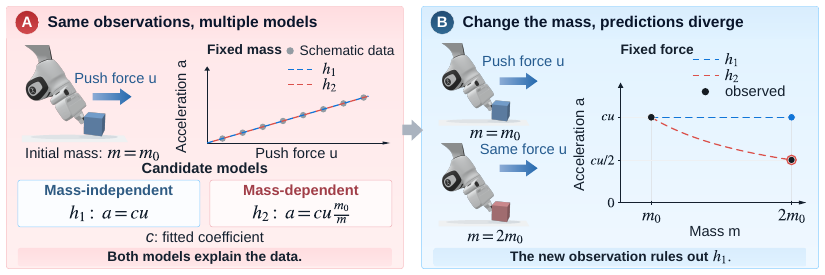}
  \caption{\lead{Different world models can fit the same trajectories;
\method selects experiments to distinguish them.} At mass $m_0$, both equations fit the observations
(A). At $2m_0$ with the same force, their predicted accelerations differ.
The observation contradicts the mass-independent prediction (B).}
  \label{fig:motivation}
\end{figure}

\lead{Agents should set up targeted experiments that actively discard incorrect models.}
For each safe and feasible experiment, the system compares predictions from
candidate world models and prioritizes experiments with larger predictive
differences~\citep{shyam2019max}. Observations can rule out models whose
predictions substantially disagree with the measured outcomes. If no candidate
explains the new observations, the existing equations may lack necessary
variables or interaction terms, providing evidence for the next revision.
Experimental outcomes thus not only add training data but help determine
whether to select an existing model or revise its equation structure.
Figure~\ref{fig:motivation} contrasts fitting existing trajectories alone with
this interactive world model learning process.

\begin{figure}[!t]
  \centering
  \includegraphics[width=\textwidth]{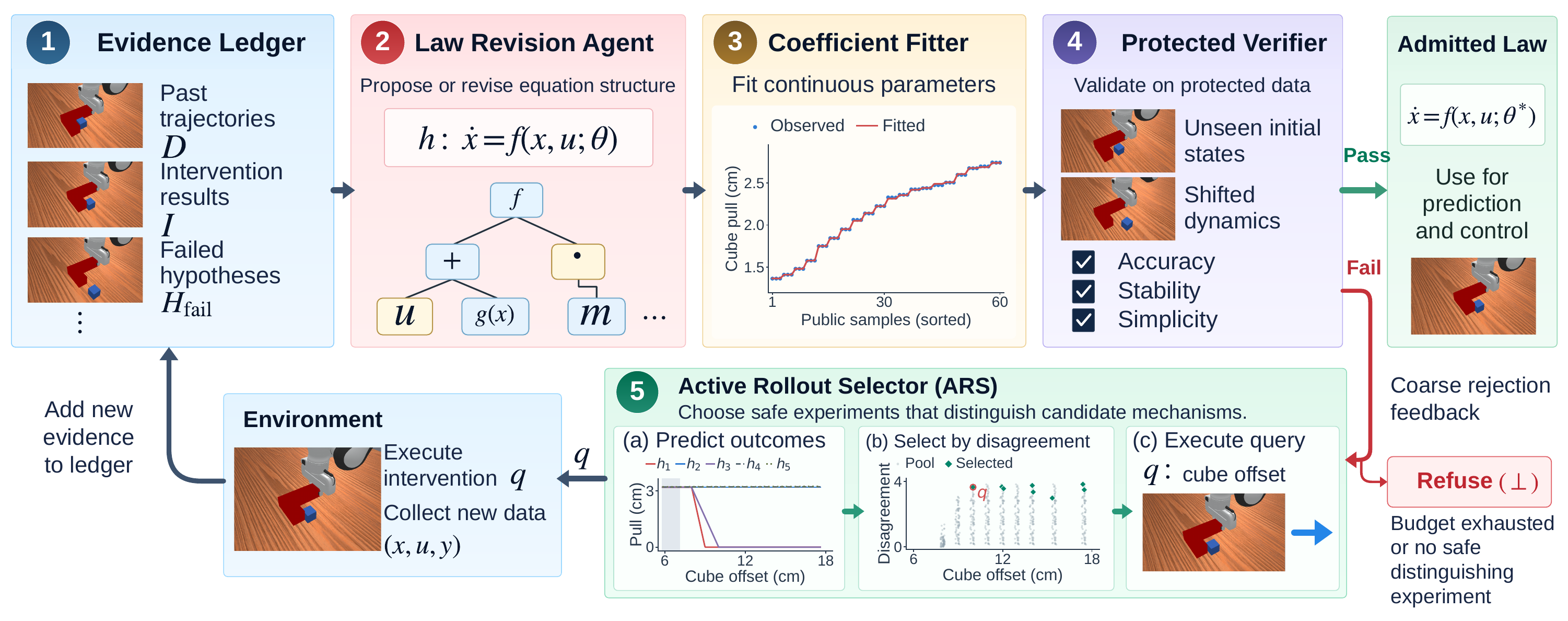}
  \caption{\lead{\method uses new observations to revise equations.}
The Law Revision Agent proposes an equation structure from public evidence,
and the Coefficient Fitter estimates its continuous parameters. The Protected
Admission Verifier checks fitted equations on held-out data; admitted equations
can be used for prediction and action selection. After rejection, ARS may
select a feasible experiment where candidate equations predict different
outcomes; its observation guides the next revision.}
  \label{fig:architecture}
\end{figure}

\lead{\method tests and revises equation-based world models.} The system initially samples a set of trajectories in a shared record (\emph{Evidence Ledger}).
It then uses this evidence to propose or revise equation structures under
type constraints (\emph{Law Revision Agent}), and a numerical optimizer then
fits their coefficients (\emph{Coefficient Fitter}). 
When multiple candidate models explain the available data, an experiment selector chooses an experiment on which their predictions disagree, subject to cost and safety constraints (\emph{Active Rollout Selector}, ARS). New observations are added
to the ledger to guide subsequent model selection and structural revision.
An independent verifier uses held-out data unavailable to the agent to decide
whether to admit a candidate model (\emph{Protected Admission Verifier}).
For robotic control, an admitted equation, the nominal simulator, and a fixed
low-level controller form a forward model that predicts the final state
resulting from an action. Action selection is the corresponding inverse
problem: given a target state, we numerically seek an action whose predicted
final state is close to the target, then execute it through the controller.
Rejected models undergo further revision, and no model is returned if none
passes validation before the interaction budget is exhausted.
Figure~\ref{fig:architecture} shows this loop and the information flow between
its components.

\lead{Our evaluation covers the full process from equation construction to
robotic application.} On custom benchmarks, \method constructs equations
beyond its initial set and uses new observations to revise incorrect models
in robotic tasks with tool-mediated pulling and non-convex object pushing. Public ODE datasets
further test its ability to model different dynamical systems. 
With candidate models and budgets fixed, the ARS module reduces the mean number of interactions needed to distinguish them. In RoboCasa drawer and door tasks
\citep{nasiriany2026robocasa365}, admitted models reduce prediction error under
distribution shift while maintaining control performance. When required
equation terms, state variables, or observations are missing, the independent
verifier rejects models unsupported by its validation data.

\lead{We use \method to study how active sampling of environment interactions improves equation-based world models and how these models can be used to support action.} Our
contributions are threefold. \textbf{(1) Experimental observations can drive
equation structure revision, not just parameter fitting.} \method connects
experiment selection with structural revision, using newly acquired observations to test existing models and iteratively construct better hypotheses. Our experiments show how this process repairs initial proposals and yields effective models beyond the initial candidate set.
\textbf{(2) Active experiments selection based on competing hypotheses disagreement can efficiently improve world-model refinement.} We fix candidate models and interaction budgets to isolate the impact of active experiments selection, finding that disagreement-based selection reduces the mean identification cost. 
\textbf{(3) Learning equation-based models supports testable prediction and naturally allows control.} In robotic tasks, our
equation-based models reduce prediction error under distribution shift. 
The explicit equations also enable action selection by solving the inverse problem, connecting model learning to goal-directed execution while maintaining control performance.

\section{Background and Related Work}

\lead{Learning an equation requires choosing its structure and fitting its
coefficients.} SINDy selects a sparse set of dynamics terms from a predefined
function library~\citep{brunton2016sindy}, whereas symbolic regression searches
expressions by combining variables and operators~\citep{cranmer2023pysr}.
LLM-SR couples language-model program proposals with numerical coefficient
fitting~\citep{shojaee2025llmsr}, while SR-Scientist uses an agent to analyze
data and revise executable formulas~\citep{xia2026srscientist}. \method
retains this separation of structure generation and coefficient fitting,
representing world models as executable equations subject to type constraints.

\lead{Active experiments can distinguish models that are difficult to separate
from existing data alone.} Active system identification selects inputs to
reduce model uncertainty~\citep{wagenmaker2020active}, APPS queries informative
phase-space regions~\citep{jiang2025apps}, and LLM-ACES selects new initial
conditions using disagreement between candidate model predictions, then uses
the new trajectories in subsequent symbolic search~\citep{abhyankar2026llmaces}.
\method applies this idea to environment interaction: it selects experiments
subject to feasibility, cost, and safety constraints, uses new observations to
revise equations, and independently validates the resulting models.

\lead{Trajectories from failed executions can inform revisions to executable
knowledge.} Kintsugi repairs an executable policy knowledge base through
constrained edits and verification~\citep{cao2026kintsugi}, while ESBM updates
symbolic behavioral models from task rollouts, adaptive questions, and
world-model probes~\citep{shindo2026esbm}. These works primarily revise skills,
predicates, and symbolic rules. \method instead revises equations describing
continuous state changes: an agent edits the equation structure, a numerical
optimizer fits coefficients, and an independent verifier decides whether to
admit the resulting model.

\FloatBarrier

\section{\method: Action-Guided Law Discovery, Evaluation, and Revision}
\label{sec:method}

\subsection{Overview and Executable Laws}

\lead{\method learns equation-based world models under a limited interaction
budget.} Its inputs are initial trajectories $\dataset_0$, a typed grammar
$\mathcal{G}$ specifying available variables, operators, and their composition
constraints, and a set of permitted environment experiments $\queries$.
Each experiment $q\in\queries$ changes an initial state, object property, or
action and returns a new trajectory $\tau(q)$. After $k$ experiments, the data
used for model learning become
\begin{equation}
 \dataset_k=\dataset_0\cup\{\tau(q_1),\ldots,\tau(q_k)\}.
 \label{eq:evidence-update}
\end{equation}
Within the proposal and interaction budgets $\mathcal{B}$, the system
constructs and tests candidate equations. It returns an independently validated
equation structure $h$, fitted coefficients $\widehat{\xi}_h$, and a stated
scope of applicability; if no candidate passes validation when the process
terminates, it refuses to return a model:
\begin{equation}
 \operatorname{ALDER}(\dataset_0,\mathcal{G},\queries,\mathcal{B})
 \longrightarrow (h,\widehat{\xi}_h)\quad\text{or}\quad\bot.
 \label{eq:alder-task}
\end{equation}
Here, $\bot$ denotes refusal to return a model.

\lead{Expression structure and numerical coefficients together define the
model's dynamics equations.} Structure $h$ is an executable expression tree
specifying how variables and operators are combined; it belongs to the set
$\hypotheses(\mathcal{G})$ permitted by the typed grammar $\mathcal{G}$.
The agent generates or revises $h$, while the Coefficient Fitter estimates
$\xi_h$ within prescribed bounds from the accumulated trajectories. Given
input $e_t$ containing task-relevant states, geometry, and actions, executing
the equation predicts the task response $y_t$:
\begin{equation}
 \widehat{y}_t=\phi_h(e_t;\xi_h),
 \qquad h\in\hypotheses(\mathcal{G}).
 \label{eq:local-program}
\end{equation}
Here, $\phi_h$ evaluates the equation specified by $h$. Depending on the task,
the predicted quantity may be acceleration, force, or a score indicating
whether an event occurs. Fixed dynamics calculations and numerical integration
convert predicted derivatives or forces into future states. This representation
supports object interactions, joint motion, and contact constraints. When
dynamics admit a pairwise decomposition, the same equation can be shared across
object pairs. The equation grammar and validation scope are detailed in
Appendices~\ref{app:grammar} and~\ref{app:law-semantics}.

\lead{\method iteratively revises candidate equations and uses independent
validation to decide whether to admit them.} The \emph{Evidence Ledger} stores
initial trajectories, experimental observations, and previous failures. The
\emph{Law Revision Agent} uses these records to propose or modify equation
structures, and the \emph{Coefficient Fitter} fits their coefficients. The
\emph{Protected Admission Verifier} evaluates candidate models on held-out data
unavailable to the agent. Following rejection, the \emph{Active Rollout Selector}
(ARS) may select a feasible experiment on which candidate models predict
different outcomes. The resulting observation is added to the ledger for the
next revision. The system returns a model that passes validation; if no candidate
passes within the budget, it refuses to return a model.

\subsection{Discovering Laws by Acting}

\lead{\method revises the equation structures using observed
trajectories and failure records.} At round $k$, the Evidence Ledger provides
variable meanings, current trajectories, fit errors and residual analyses,
executed experiments, and permitted rejection categories. In an isolated
sandbox, the Law Revision Agent reads these records and the typed grammar
$\mathcal{G}$ to produce a candidate structure $h_k$ as a JSON expression tree,
together with parameter names and bounds, an evidence-based rationale, and an
intended scope. Only candidates passing grammar, type, expression-size, and
executability checks proceed to coefficient fitting and validation by the
subsequent components. The agent proposes structures; it neither fits
coefficients nor decides whether a model is admitted.

\lead{Once the structure is fixed, the Coefficient Fitter estimates only
numerical coefficients.} We extract $N_k$ input--target pairs $(e_n,y_n)$ from
the current trajectories $\dataset_k$ and minimize prediction loss over the
permitted coefficient domain $\Xi_{h_k}$:
\begin{equation}
 \widehat{\xi}_{h_k}
 =\arg\min_{\xi\in \Xi_{h_k}}
 \frac{1}{N_k}
 \sum_{n=1}^{N_k}
 \ell\!\left(\phi_{h_k}(e_n;\xi),y_n\right).
 \label{eq:fit}
\end{equation}
We use squared error for continuous targets and class-balanced logistic loss
for events. Squared-error models linear in the coefficients use ridge
regression; nonlinear parameters use bounded reparameterization and gradient
optimization (Appendix~\ref{app:procedure}). For ridge regression,
Equation~\ref{eq:fit} omits the regularization term for brevity.
Fitting updates only coefficients;
missing variables, operators, or compositions require the agent to revise the
equation structure.

\lead{The Active Rollout Selector (ARS) uses prediction disagreement to choose
experiments while accounting for safety and cost.} At round $k$, the candidate
set contains equation structures, fitted coefficients, and weights:
\begin{equation}
 \mathcal{C}_k=
 \left\{(h_m,\widehat{\xi}_m,w_m)\right\}_{m=1}^{M_k},
 \qquad \sum_{m=1}^{M_k}w_m=1.
 \label{eq:comparison-set}
\end{equation}
Candidates come from current or earlier proposals, grammar-permitted
alternatives, or a supplied set. All are refitted on the same $\dataset_k$ and
weighted by public validation likelihood, error, or BIC; weights are uniform
when no calibrated score is available. The set need not contain the true
equation. Neither protected data nor actual outcomes of unqueried experiments
inform its construction or experiment selection. For each feasible experiment
$q$, candidates predict trajectories $\widehat{\tau}_{h_m}(q)$, and ARS selects
\begin{equation}
 q^\star=\arg\max_{q\in\queries:S(q)=1}
 \left[
 \sum_{a<b}\widetilde{w}_{ab}
 d\!\left(\widehat{\tau}_{h_a}(q),
           \widehat{\tau}_{h_b}(q)\right)
 +\lambda_D D(q,\dataset_k)-\lambda_c c(q)
 \right],
 \label{eq:query}
\end{equation}
where $d$ measures prediction disagreement, $\widetilde{w}_{ab}$ emphasizes
plausible model pairs, $D$ rewards experimental inputs unlike those already
observed, $c$ measures interaction cost, and $S(q)=1$ denotes satisfaction of
safety constraints. A rejected proposal can also be compared directly with
alternative models to guide further experiments. Prediction disagreement
serves as a proxy for information gain, not an exact mutual-information
calculation.

\lead{The novel experiments allow to discriminate candidate models and inform equation revision.} The environment executes the selected experiment
$q^\star$ and adds its trajectory to the dataset through
Equation~\ref{eq:evidence-update}. An observation that supports some candidates
and contradicts others helps distinguish models; if no candidate explains it,
the equation structures may need revision. The Evidence Ledger records failed
predictions, the corresponding state--action conditions, and public residual
diagnostics. Using this evidence, the Law Revision Agent may add or remove
variables, change operators, or revise compositions and piecewise branches
within $\mathcal{G}$. After each edit, the Coefficient Fitter refits coefficients
from scratch, and the independent verifier decides whether to admit the
candidate model.

\subsection{Independent Admission and Downstream Use}

\lead{Candidate models are admitted only after passing independent
validation.} The Protected Admission Verifier uses protected data unavailable
to the agent to check prediction error, improvement over a simple reference,
expression complexity, numerical stability, finite outputs, and performance
under task-specific distribution shifts. Separate public splits support
coefficient fitting and diagnostics; final test data do not participate in
model construction, experiment selection, or admission. After rejection, the
agent receives only public diagnostics and a coarse failure category, never
protected data or scores, the target equation, or final test outcomes. A
candidate passing all preset criteria is returned as $(h,\widehat{\xi}_h)$.
Otherwise, revision continues until a candidate passes, no distinguishing safe
experiment remains, or the proposal or interaction budget is exhausted; the
latter two cases return $\bot$.

\lead{A validated world model supports control as an inverse problem.} Given
state $X_0$, the learned joint-force equation, nominal simulator, and fixed
low-level controller jointly predict the joint position $\widehat{q}_T(u)$
after executing action $u$ for $T$ steps. For target position $g$, control seeks
an action satisfying
\begin{equation}
 \left|\widehat{q}_T(u^\star)-g\right|\leq\varepsilon.
 \label{eq:planning}
\end{equation}
With the equation and coefficients fixed, numerical root finding searches
bounded actions over preset horizons. Within the model-evaluation budget, we
return an action meeting target tolerance $\varepsilon$ and predicted safety
constraints; otherwise, we return the closest-to-target evaluated action
satisfying the safety constraints. Without a validated model or an action
satisfying these safety constraints, execution is refused. Control success is
measured from actual execution.

\FloatBarrier

\section{Experiments}
\label{sec:experiments}

\begin{table}[!t]
\caption{\lead{Interactive revision improves functional recovery; fitted
equations predict ODE trajectories under shift.}
A: NovelLaw reports functional recovery (\%) and median shifted NMSE over
60 cases; robot shifted NMSE is mean $\pm$ SD over five
episode-disjoint partitions. Rejected runs are included.
B: median trajectory NMSE and mean symbolic accuracy on 122 ODE systems.}
\label{tab:rq1}
\centering
\small
\setlength{\tabcolsep}{2.5pt}
\renewcommand{\arraystretch}{0.95}
\begin{tabularx}{\linewidth}{@{}>{\raggedright\arraybackslash}Xrrrrrr@{}}
\toprule
\multicolumn{7}{@{}l@{}}{\textbf{A. Controlled discovery}} \\
& \multicolumn{2}{c}{NovelLaw (60 cases)}
& \multicolumn{2}{c}{PullCubeTool (5 runs)}
& \multicolumn{2}{c}{PushT (5 runs)} \\
\cmidrule(lr){2-3}\cmidrule(lr){4-5}\cmidrule(lr){6-7}
Method & \shortstack{Functional\\recovery (\%, $\uparrow$)}
& \shortstack{Shift\\NMSE ($\downarrow$)}
& \shortstack{Admitted\\($\uparrow$)}
& \shortstack{Shift\\NMSE ($\downarrow$)}
& \shortstack{Admitted\\($\uparrow$)}
& \shortstack{Shift\\NMSE ($\downarrow$)} \\
\midrule
One-shot agent & 43.3 & $6.54\!\times\!10^{-2}$ & 0/5 & $4.816\pm1.973$ & 3/5 & $0.842\pm0.644$ \\
Revision-only agent & 75.0 & $1.50\!\times\!10^{-2}$ & 0/5 & $1.296\pm0.024$ & 3/5 & $1.292\pm1.638$ \\
Random-query agent & \textbf{85.0} & $9.15\!\times\!10^{-3}$ & \textbf{5/5} & $0.194\pm0.138$ & \textbf{4/5} & $0.673\pm0.473$ \\
\rowcolor{alderrow}
\textbf{\method} & \textbf{85.0} & $\mathbf{5.58\!\times\!10^{-3}}$ & \textbf{5/5} & $\mathbf{0.161\pm0.052}$ & \textbf{4/5} & $\mathbf{0.420\pm0.185}$ \\
\midrule
\end{tabularx}
\par
\begin{tabularx}{\linewidth}{@{}>{\footnotesize\raggedright\arraybackslash}Xrrrr@{}}
\multicolumn{5}{@{}l@{}}{\textbf{B. ODEBench + ODEBase (122 systems)}} \\
Method & \makebox[6.4em][r]{\shortstack{Reconstruction\\NMSE ($\downarrow$)}}
& \makebox[6.4em][r]{\shortstack{Generalization\\NMSE ($\downarrow$)}}
& \makebox[6.4em][r]{\shortstack{Future-time\\NMSE ($\downarrow$)}}
& \shortstack{Symbolic\\accuracy (\%, $\uparrow$)} \\
\midrule
SINDy~\citep{brunton2016sindy} & 2.18e-14 & 1.55e-04 & 2.68e-05 & 15.6 \\
PySR~\citep{cranmer2023pysr} & 1.91e-05 & 1.46e-02 & 1.38e-01 & 9.0 \\
QBC~\citep{abhyankar2026llmaces} & 3.01e-05 & 1.77e-03 & 6.43e-02 & 15.6 \\
BO~\citep{abhyankar2026llmaces} & 2.73e-05 & 1.02e-03 & 4.30e-02 & 18.9 \\
SR-Scientist~\citep{xia2026srscientist} & 1.39e-13 & 2.08e-12 & 8.88e-11 & 40.2 \\
LLM-ACES~\citep{abhyankar2026llmaces} & 3.02e-12 & 2.01e-04 & 3.70e-05 & 24.6 \\
\rowcolor{alderrow}
\textbf{\method} & \textbf{2.26e-15} & \textbf{6.01e-16} & \textbf{1.61e-29} & \textbf{86.9} \\
\bottomrule
\end{tabularx}
\end{table}

\lead{Reliable world model learning must address accurate model construction,
interaction efficiency, reliable validation and can allow control utility.} We evaluate \method using four research questions.
\textbf{RQ1:} Can \method construct
equations beyond its initial set and use failure evidence to revise candidate
models? \textbf{RQ2:} With candidate models and interaction budgets fixed, can
ARS identify the target model with fewer interactions? \textbf{RQ3:} Do
validated models improve prediction under distribution shift and support
control in RoboCasa? \textbf{RQ4:} When equation terms, state variables, or
observations are insufficient, does the system refuse models unsupported by
the available evidence?

\textbf{Experimental setup.} Except for oracle references with privileged
information, methods access only public trajectories and permitted intervention
outcomes, not hidden environment ground truth, protected admission data, or
final test data. Controlled comparisons use the same splits and match
observations, query pools, coefficient fitters, and interaction budgets
according to the comparison's purpose. Invalid proposals and refusals remain
in the denominators for model recovery and control success. Full protocols are
provided in Appendix~\ref{app:details}.

\subsection{ALDER Constructs and Repairs Laws Beyond the Initial Set (RQ1)}

\lead{To evaluate how well models discover and update dynamical equations, we test them on both custom interaction tasks and public benchmark systems.} Our \emph{NovelLaw} benchmark evaluates out-of-set equation construction, while custom \emph{PullCubeTool} and \emph{PushT} protocols in ManiSkill \citep{tao2025maniskill3} test whether replayed actions over varied poses drive revision of contact-dependent dynamics. Finally, we evaluate equation discovery on 122 public dynamical systems: 63 from ODEBench~\citep{dascoli2024odeformer} and 59 from ODEBase~\citep{lueders2022odebase}. See Appendix~\ref{app:external-ode} for the evaluation protocol and additional results.

\lead{We evaluate baseline variants to isolate the individual contributions of structural revision and interactive querying.} 
One-shot acts as a single-proposal reference, revision-only tests iterative editing without new data, and random-query shares \method's interaction budget but selects queries uniformly at random. 
All variants use identical grammars, coefficient fitters, and admission criteria. 
Table~\ref{tab:rq1-full} adds fixed-library, SINDy, and MLP baselines for comparison against fixed structures, sparse regression, and neural networks. On the 122 ODE systems, we compare \method against sparse, symbolic, and active discovery baselines under a common evaluation protocol (Table~\ref{tab:rq1}B; Appendix~\ref{app:external-ode}).

\begin{figure}[!t]
\centering
\includegraphics[width=\linewidth]{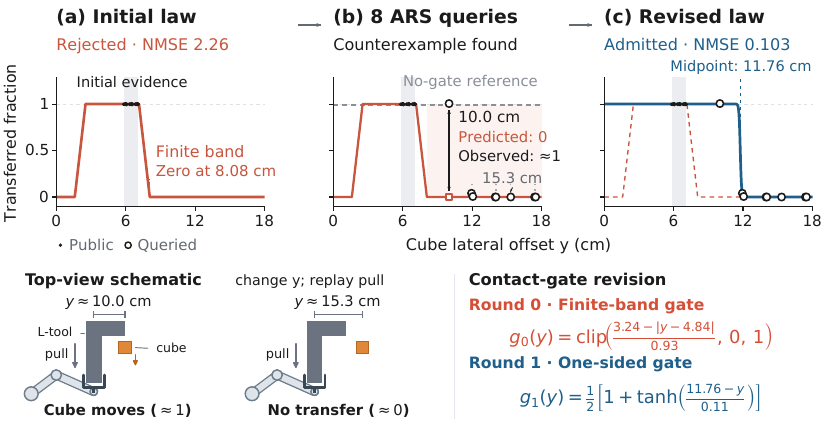}
\caption{\lead{An ARS query reveals a failed prediction and guides revision
of the PullCubeTool contact gate.} (a) The initial equation fits demonstrations
but fails the admission checks. (b) At 10\,cm, the observed transfer fraction
is near one, but the equation predicts zero. (c) The equation with a revised
one-sided gate is admitted. Filled and hollow points mark public and queried
observations; some query points overlap. The schematics show the same pull
at two cube positions; the formulas describe only the contact gates. NMSE is
measured on the protected admission set.}
\label{fig:rq1-example}
\end{figure}

\lead{Interaction evidence helps \method revise initial equations and improve
prediction under distribution shift.} On NovelLaw, \method retains every
initial success and increases recovery from 26/60 to 51/60, matching
random-query's recovery with lower shifted NMSE. On PullCubeTool, all \method
runs pass predictive validation, whereas one-shot and revision-only never
pass. Across both robot tasks, \method matches random-query's admission counts
but has lower mean shifted error (Table~\ref{tab:rq1}A).
Figure~\ref{fig:rq1-example} shows how a new observation contradicts the
initial prediction and prompts revision of the contact gate from a finite
band to a one-sided structure. Across the 122 ODE systems, \method achieves
the lowest median NMSE in all three prediction settings and the highest
symbolic accuracy among the methods shown in Table~\ref{tab:rq1}B. It achieves
functional recovery on 106/122 systems (86.9\%), with trajectory NMSE at most
$10^{-6}$ in reconstruction, generalization to unseen initial conditions, and
continuous future-time extrapolation. Under the common symbolic evaluation,
all equations match in 106/122 systems (86.9\%).

\subsection{ARS Identifies Mechanisms Faster than Random Queries (RQ2)}

\lead{We compare experiment selectors with fixed candidate models and equal
eight-query budgets.} Across 182 initially ambiguous cases, methods share
initial observations, feasible query pools, and noise settings, refitting
coefficients without changing equation structures. We compare ARS with passive
sampling, random selection, geometric coverage, and parameter uncertainty,
alongside an oracle reference. All trials complete eight queries; costs are
computed retrospectively from sustained correct identification, with unresolved
cases assigned a cost of nine. Appendix~\ref{app:intervention-selection}
details the mechanism families, protocols, and paired statistical comparisons.

\begin{figure}[!t]
\centering
\includegraphics[width=\linewidth]{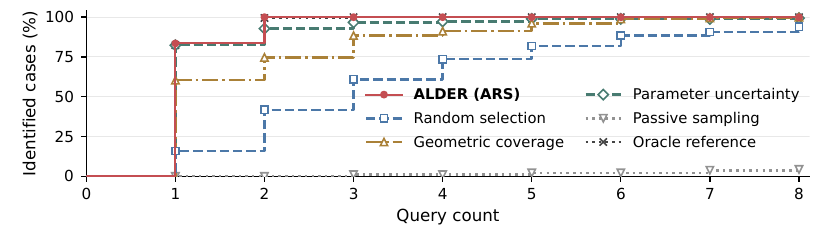}
\caption{\lead{In complete eight-query records, ARS identifies the target
model earlier than random selection.} Curves show the cumulative fraction
reaching sustained correct identification among fixed candidate models,
computed retrospectively. Unresolved cases remain in the denominator.}
\label{fig:rq2}
\end{figure}

\lead{Active queries driven by mechanism disagreement consistently minimize identification cost across all tested baselines.} \method identifies
every target with a median identification cost of one, whereas random selection
has a median cost of three and leaves some targets unresolved
(Figure~\ref{fig:rq2}).
Their mean penalized identification costs are 1.16 and 3.53, respectively.
Geometric coverage also identifies every target and, like parameter
uncertainty, has a median cost of one. Their mean costs, however, are 1.91
and 1.35, both above \method. Paired statistics support these mean-cost
advantages over both strong baselines, showing that prediction disagreement
can further improve identification efficiency when the candidate models
are given in advance (Appendix~\ref{app:intervention-selection}).

\subsection{Explicit Laws Improve Shifted Prediction and Preserve Control (RQ3)}

\lead{We test world models for prediction and control in RoboCasa's
\texttt{OpenDrawer} and \texttt{OpenDoor} tasks}
\citep{nasiriany2026robocasa365,zhu2020robosuite}.
A fixed controller first grasps the handle; the validated world model then
predicts pull outcomes, and numerical inversion selects pull amplitude and
duration for a target joint position.
We vary joint damping, Coulomb friction, spring stiffness, and their composition
in MuJoCo \citep{todorov2012mujoco} to test performance under changing dynamics.
Each task combines four dynamics families with ten random seeds, yielding
40 simulated worlds. Models use simulator-provided state and force observations.

\begin{table}[!htbp]
\caption{\lead{Explicit dynamics discovered by \method improve shifted
prediction while preserving downstream control performance in RoboCasa.}
Prediction RMSE and control MAE are in $10^{-3}$ normalized joint-position
units; success rates are percentages. Values are mean $\pm$ SD across ten
seeds, each averaging four mechanism families. Control uses 640 targets per
model and task; bold denotes the best reported performance.}
\label{tab:robocasa-control}
\centering
\small
\setlength{\tabcolsep}{2.8pt}
\renewcommand{\arraystretch}{1.08}
\begin{tabular}{@{}lrrrrrr@{}}
\toprule
& \multicolumn{3}{c}{OpenDrawer} & \multicolumn{3}{c}{OpenDoor} \\
\cmidrule(lr){2-4}\cmidrule(lr){5-7}
Model & \shortstack{OOD prediction\\RMSE ($\downarrow$)}
& \shortstack{Control\\MAE ($\downarrow$)}
& \shortstack{Success\\(\%, $\uparrow$)}
& \shortstack{OOD prediction\\RMSE ($\downarrow$)}
& \shortstack{Control\\MAE ($\downarrow$)}
& \shortstack{Success\\(\%, $\uparrow$)} \\
\midrule
Nominal & $6.8\pm0.7$ & $5.04\pm0.54$ & $83.6\pm2.2$ & $7.2\pm0.8$ & $4.26\pm0.37$ & $94.4\pm2.0$ \\
Passive law & $7.5\pm1.0$ & $5.78\pm0.76$ & $80.3\pm3.6$ & $7.4\pm1.5$ & $3.80\pm0.49$ & $89.7\pm3.1$ \\
Matched MLP & $3.7\pm0.8$ & $1.98\pm0.35$ & $96.9\pm2.4$ & $3.1\pm0.7$ & $1.81\pm0.32$ & $99.5\pm1.1$ \\
\rowcolor{alderrow}
\textbf{ALDER} & $\mathbf{0.9\pm0.1}$ & $\mathbf{0.74\pm0.10}$ & $\mathbf{100.0\pm0.0}$ & $\mathbf{0.9\pm0.1}$ & $\mathbf{0.49\pm0.06}$ & $\mathbf{100.0\pm0.0}$ \\
\bottomrule
\end{tabular}

\end{table}

\lead{We use a common numerical inverse solver to test the effects of
interactive revision and model representation.} The nominal model does not
adapt to dynamics changes and provides an initial reference. The passive law
uses only initial trajectories, testing the joint benefit of new interactions
and subsequent revision. The matched MLP and \method share training data,
action ranges, and solver budgets to compare neural and equation-based models.
We report shifted prediction error, actual endpoint-to-target error, and
control success. Success requires safe actual execution and normalized
target-position error at most 0.01. Full settings and supplementary results
appear in Appendix~\ref{app:robocasa-protocol}.

\lead{Across both tasks, validated equation-based models reduce prediction
and control errors while maintaining high target-reaching success.} With the
same training data and numerical solver, \method predicts more accurately
under distribution shift than the matched MLP, and its selected actions yield
endpoints closer to the targets. Both achieve high control success, while
\method meets the success criteria on all tested targets
(Table~\ref{tab:robocasa-control}). Relative to the passive law trained only on
initial trajectories, \method improves prediction accuracy, control accuracy,
and success, supporting the joint benefit of new interaction evidence and
subsequent equation revision.

\subsection{Verification Rejects Laws Unsupported by Available Evidence (RQ4)}

\lead{We separately restrict equation structure and available observations
to test rejection of unsupported world models.} With the library fixed, we
compare complete models, missing global smooth terms, missing local contact
terms, and hidden key state over 20 seeds per regime, using an auxiliary
residual diagnostic to locate states where errors concentrate. In RoboCasa's
Drawer and Door tasks, we add noise to force observations or estimate force
from position, velocity, and known control actions to assess how limited
observations affect admission, prediction coverage, and control success.
Control uses the same numerical
inverse solver and success criteria as RQ3. Full settings appear in
Appendices~\ref{app:boundary-protocol} and~\ref{app:robocasa-observation}.

\begin{table}[!t]
\caption{\lead{Admission checks reject models with missing equation terms or
state variables; degraded force observations reduce prediction coverage.}
A: AUROC and gated-residual NMSE are evaluated under distribution shift and
do not determine admission. B: Drawer and Door are pooled; coverage is the
fraction of worlds with admitted equations. RMSE uses those worlds only,
whereas control success covers all targets, counting refusals as failures.
Values with $\pm$ are mean $\pm$ SD.}
\label{tab:rq4}
\centering
\small
\setlength{\tabcolsep}{1.5pt}
\renewcommand{\arraystretch}{1.08}
\begin{minipage}[t]{0.523\linewidth}
\centering
\begin{tabular}[t]{@{}lrrr@{}}
\toprule
\multicolumn{4}{l}{\textbf{A. Missing structure or state}} \\
Regime & \shortstack{Admitted\\models}
& \shortstack{Localization\\AUROC ($\uparrow$)}
& \shortstack{Gated NMSE\\($\times10^{-3}$, $\downarrow$)} \\
\midrule
Complete & 100.0\% & \multicolumn{1}{c}{--} & $0.042\pm0.029$ \\
Missing smooth & 0.0\% & $0.494\pm0.019$ & $14.5\pm3.0$ \\
Missing local & 0.0\% & $0.969\pm0.007$ & $18.5\pm2.9$ \\
Hidden local & 0.0\% & $0.997\pm0.001$ & $168\pm114$ \\
\bottomrule
\end{tabular}
\end{minipage}\hfill
\begin{minipage}[t]{0.465\linewidth}
\centering
\begin{tabular}[t]{@{}lrrr@{}}
\toprule
\multicolumn{4}{l}{\textbf{B. Robot observation channel}} \\
Channel & \shortstack{Pred.\\coverage}
& \shortstack{RMSE\\($\times10^{-3}$, $\downarrow$)}
& \shortstack{Control\\success ($\uparrow$)} \\
\midrule
Force reference & 100.0\% & $0.9\pm0.7$ & 100.0\% \\
Force noise 1\% & 80.0\% & $1.0\pm0.7$ & 79.9\% \\
Force noise 5\% & 58.8\% & $1.0\pm0.7$ & 58.8\% \\
Force noise 10\% & 42.5\% & $1.2\pm0.7$ & 42.5\% \\
Estimated force & 0.0\% & \multicolumn{1}{c}{--} & 0.0\% \\
\bottomrule
\end{tabular}
\end{minipage}
\end{table}
\lead{Under the tested conditions, \method rejects incomplete models and limits
admission as observation quality degrades.} All complete models pass
verification. For missing local terms or key state, all models are refused and
auxiliary localization is effective. With a missing global smooth term, all
models are still refused despite near-chance localization under distribution
shift (Table~\ref{tab:rq4}A). In the robot tasks, higher force-observation noise
reduces prediction coverage; models fitted to the current motion-based force
estimates do not pass validation (Table~\ref{tab:rq4}B). Prediction error remains low
among admitted models, whereas control success is computed over all targets,
with refusals and unsuccessful executions both counted as failures.
\FloatBarrier

\section{Limitations and Future Work}
\label{sec:limitations}

\lead{Future work will expand \method's model expressiveness and robot
applications while improving learning and verification reliability.}
The current method uses a given state representation and mathematical grammar,
with experiments mainly based on short simulated interactions and state and
force observations. Extensions can expand equation operators and intervention
interfaces, incorporate visual perception and contact trajectory information,
and evaluate perception error, interaction cost, and safety on physical robots.
Candidate models may share prediction errors, and reusing validation data may
lead to overfitting. Increasing candidate diversity, limiting verification
feedback, and testing on independent data offer directions for improving
reliability and studying theoretical guarantees.
\section{Conclusion}

\lead{We introduce \method to actively test and revise equation-based world
models through environment interaction.} Prediction disagreement among candidate
models guides experiment selection, interaction counterevidence drives revision
of equation structure, and independent verification determines whether models
are admitted for prediction and control. Experiments show that \method constructs
equations beyond its initial set, identifies target models earlier among given
candidates, improves shifted prediction while preserving robot control
performance, and refuses unsupported models.
Environment interaction thus provides not only training data but also a tool
for identifying, falsifying, and revising world models.

\bibliography{references}
\bibliographystyle{iclr2027_conference}

\clearpage
\appendix
\section{Implementation and Evaluation Details}
\label{app:details}
\suppressfloats[t]

\lead{This appendix follows the implementation and four main evaluations,
then presents additional validation and reproducibility details.}
Appendix~\ref{app:implementation} specifies the equation-based model implementation.
Appendices~\ref{app:law-construction}--\ref{app:verification} cover construction,
identification, control, and verification.
Appendix~\ref{app:extensions} reports additional systems and proposal backbones,
and Appendix~\ref{app:audit} reports statistical protocols, model settings,
and resource use.

\subsection{ALDER Separates Proposal, Fitting, and Verification}
\label{app:implementation}

\subsubsection{Meaning and Scope of an Executable Equation}
\label{app:law-semantics}

\lead{An executable equation combines a structure, fitted coefficients, and a
stated prediction scope.} The fitted pair $(h,\widehat{\xi}_h)$ defines this
equation over its stated inputs and outputs. The typed grammar $\mathcal{G}$ specifies the
available variables and operators; $h\in\hypotheses(\mathcal{G})$ selects a
permitted equation structure, and $\widehat{\xi}_h$ contains its fitted
numerical coefficients. For task input $e_t$, the program
$\phi_h(e_t;\widehat{\xi}_h)$ predicts the task-defined response $y_t$.
Admission means that this fitted program passes the predeclared checks on
harness-held validation data, including applicable shifted regimes. These
checks support prediction in the tested scope; they do not establish that the
program uniquely describes the underlying physical dynamics. Recovery against a known
reference equation is assessed separately, after discovery is complete.

\subsubsection{Procedure and Coefficient Fitting}
\label{app:procedure}

\lead{The experiment harness separates proposal, fitting, experiment selection,
and protected verification throughout the discovery loop.}
Here, the harness is the software that runs the discovery procedure, distinct
from the low-level robot controller used for execution.
Algorithm~\ref{alg:discovery} gives the complete procedure. The grammar, split
roles, proposal and query budgets, admission thresholds, and final evaluation
seeds are frozen before discovery begins.

\begin{algorithm}[H]
\caption{\method's Verified Law Discovery Loop}
\label{alg:discovery}
\small
\begin{algorithmic}[1]
\algrenewcommand{\algorithmicrequire}{\textbf{Input:}}
\algrenewcommand{\algorithmicensure}{\textbf{Output:}}
\Require Public evidence $\dataset_0$, grammar $\mathcal{G}$, feasible experiments
$\queries$, and budgets $\mathcal{B}$.
\Require Protected admission and shifted validation data $\mathcal{V}$.
\Ensure A validated model $(h,\widehat{\xi}_h)$, or refusal $\bot$.
\State Initialize the \emph{Evidence Ledger} $\mathcal{E}_0$ from $\dataset_0$;
set $k\gets 0$.
\State Create an isolated repository containing only public evidence,
grammar,\newline instructions, proposal schema, and permitted failure records.
\While{proposal budget remains}
  \State $h_k\gets$ \emph{Law Revision Agent}$(\mathcal{E}_k,\mathcal{G})$;
  charge the proposal budget.
  \If{schema, arity, variable, bound, size, or execution checks fail}
    \State Record the public validation failure in $\mathcal{E}_k$;
    \textbf{continue}.
  \EndIf
  \State Compile $h_k$; fit $\widehat{\xi}_{h_k}$ with the
  \emph{Coefficient Fitter} on $\dataset_k$ only.
  \State Record public fit diagnostics in $\mathcal{E}_k$.
  \State $(a_k,r_k)\gets$ \emph{Protected Admission Verifier}
  $(h_k,\widehat{\xi}_{h_k};\mathcal{V})$.
  \If{$a_k=\mathrm{accept}$}
    \State Freeze $(h_k,\widehat{\xi}_{h_k})$ and $\mathcal{E}_k$;
    \Return $(h_k,\widehat{\xi}_{h_k})$.
  \EndIf
  \State Add only the permitted rejection category $r_k$ to $\mathcal{E}_k$.
  \State \textbf{if} proposal or query budget is exhausted \textbf{then break}.
  \State Build and refit candidate models $\mathcal{C}_k$ using public
  evidence only.
  \State $q^\star\gets$ \emph{Active Rollout Selector}
  $(\mathcal{C}_k,\dataset_k,\queries)$\newline
  within the remaining budget, without accessing unrevealed outcomes.
  \State \textbf{if} no safe distinguishing experiment is available
  \textbf{then break}.
  \State Commit $q^\star$, obtain its outcome $\tau(q^\star)$, and charge the
  query budget.
  \State $\dataset_{k+1}\gets\dataset_k\cup\{\tau(q^\star)\}$.
  \State $\mathcal{E}_{k+1}\gets\operatorname{Update}(\mathcal{E}_k,
  \tau(q^\star),\text{public failure diagnostics})$; $k\gets k+1$.
\EndWhile
\State Freeze the unresolved ledger; \Return $\bot$.
\end{algorithmic}
\end{algorithm}

\lead{Recovery against reference equations is evaluated outside the discovery loop.}
After the program or unresolved ledger is frozen, recovery is
evaluated once on evaluator-only points. These points never enter proposal,
coefficient fitting, experiment selection, or admission.

\lead{Coefficient fitting optimizes numerical parameters without changing
program structure.} For squared-error objectives linear in the coefficients,
the fitter uses a ridge-regularized linear solve. For nonlinear parameters,
each bounded coefficient $\xi_r\in[a_r,b_r]$ is represented as
\begin{equation}
 \xi_r=a_r+(b_r-a_r)\,\sigma(\rho_r),
 \label{eq:bounded-parameter}
\end{equation}
where $\sigma$ is the logistic sigmoid. The unconstrained $\rho_r$ is updated
by backpropagation with multiple restarts and gradient clipping. This
reparameterization enforces the proposal's declared coefficient bounds during
optimization.

\subsubsection{Typed Formula Grammar}
\label{app:grammar}

\begin{table*}[t]
\caption{\lead{Typed grammars expose building blocks, not complete target
equations.} In RQ1, the variables and primitive
operators are available, but complete target expressions and harness-held
reference bases are absent from the proposal sandbox.}
\label{tab:grammar}
\centering
\scriptsize
\setlength{\tabcolsep}{3pt}
\begin{tabular}{p{1.9cm}p{4.2cm}p{4.2cm}p{1.9cm}}
\toprule
Track & Variables & Additional operators & Output / limit \\
\midrule
NovelLaw & distance, radial speed, masses, radii, radial action & protected division, literal power, exponential, $\tanh$ & acceleration; 31 nodes \\
PullCubeTool & tool-frame cube pose, tool/cube velocities, relative axis, commanded pull & protected division, power, exponential, trigonometric functions, absolute value, maximum & pull-axis displacement; 95 nodes \\
PushT & T-frame pusher pose, no-object TCP displacement, swept-contact indicator & protected division, power, exponential, trigonometric functions, absolute value, maximum & angular displacement; 95 nodes \\
Door & current and initial hinge/latch angles & literal power, sine, cosine & displacement; 127 nodes \\
Peg & hole-frame $x,y,z$ and hole radius & literal power, absolute value, pairwise maximum & signed score; 63 nodes \\
\bottomrule
\end{tabular}
\end{table*}

\lead{Every typed grammar combines shared safe arithmetic with a
benchmark-specific operator subset.} All grammars contain variables, finite constants, named bounded
parameters, addition, subtraction, and multiplication. Protected division
floors denominator magnitude at $10^{-6}$, exponentials are clipped, and powers
receive a positive base floor. The harness independently validates, compiles,
and refits every submitted program.

\lead{Pairwise decomposition is one instance of the equation-based model
representation.} Inputs to Equation~\ref{eq:local-program} may include relative
object states, joint position and velocity, geometric margins, or applied
actions. When a multi-object system admits a pairwise decomposition, the same
local program is shared across object pairs:
\begin{equation}
 \widehat{\dot{x}}_t^{\,i}
 =f_{\mathrm{self}}(x_t^i,u_t)
 +\sum_{j\neq i}\phi_h(e_{ij,t};\xi_h).
 \label{eq:law}
\end{equation}
Here, $f_{\mathrm{self}}$ describes the object's self dynamics and $e_{ij,t}$
contains pairwise features. This decomposition is not required for articulated
systems or contact constraints represented directly by
Equation~\ref{eq:local-program}.

\lead{Fixed-observation grammar ablations test the influence of available
physical terms on discovery.} Across all 80 RoboCasa worlds, we fix the public
observations acquired by \method and vary only the permitted equation terms.
The four conditions use the original grammar, add ten candidate terms, remove
the linear-position term, or remove both friction-sign terms. Each condition
allows at most three structural proposals per world, with the final candidate
selected using public validation data.

\begin{table}[H]
\caption{\lead{Removing key physical terms reduces admission coverage.}
All conditions use the same public observations across 40 Drawer and 40 Door
worlds.}
\label{tab:grammar-sensitivity}
\centering
\small
\begin{tabular}{lrrr}
\toprule
Grammar condition & Drawer admitted & Door admitted & Total \\
\midrule
Original grammar & 35/40 & 31/40 & 66/80 \\
Add ten candidate terms & 37/40 & 30/40 & 67/80 \\
Remove linear-position term & 28/40 & 20/40 & 48/80 \\
Remove both friction-sign terms & 18/40 & 13/40 & 31/80 \\
\bottomrule
\end{tabular}
\end{table}

\lead{Removing key physical terms reduces admission coverage, while adding
the tested candidate terms leaves coverage largely unchanged.} Removing the
linear-position term retains other position transforms, and the additional
candidate terms include correlated or redundant transforms.

\subsubsection{Sandbox and Leakage Controls}

\lead{Each NovelLaw proposal runs in a sandbox that excludes protected
evidence.} Every case uses a separate nested Git repository. The proposal agent
sees a Bubblewrap namespace in which the parent project is absent, inherited
environment variables are cleared, network use is disallowed by task contract,
and temporary authentication material is removed after session startup. The
target expression, generator seed, active-pool outcomes, admission data, and
shift data are never materialized inside the sandbox. The primary final audit
covers 60 cases, six reported methods per case, 125 agent rounds, 65 active
rounds, and 1,560 selected active rows. A unit test verifies that changing
hidden pool outcomes cannot change the selected query indices.

\FloatBarrier

\subsection{ALDER Constructs and Revises Equations Beyond Its Initial Set}
\label{app:law-construction}

\subsubsection{NovelLaw Protocol}

\begin{figure}[!t]
\centering
\includegraphics[width=\textwidth]{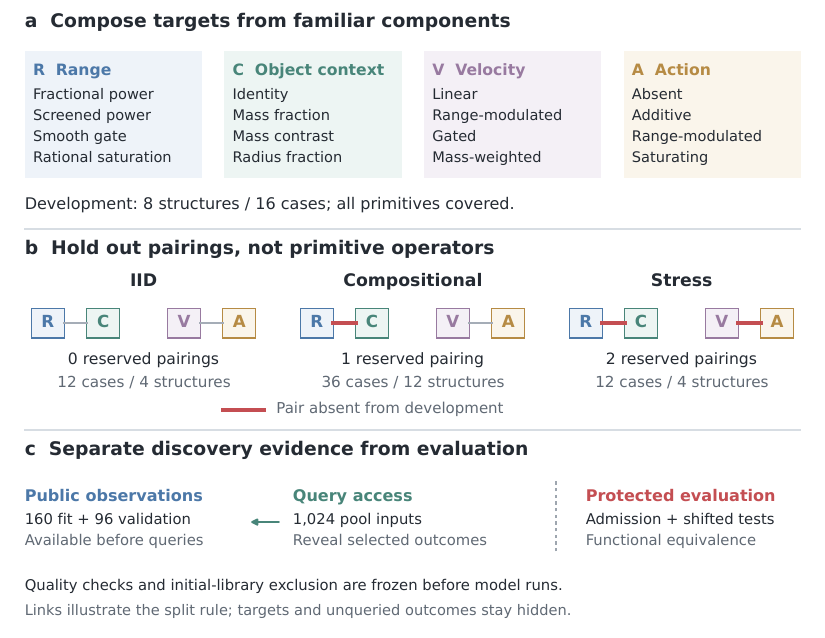}
\caption{\lead{NovelLaw tests unseen pairings of familiar equation components.}
(a) Four component families define the target structures.
(b) Red links schematically mark pairings absent from benchmark development;
they are not equation-tree edges. (c) Public observations and queried outcomes
support discovery, while protected sets evaluate admission, shift, and
functional equivalence. Per-case public validation is distinct from the
16 benchmark development cases.}
\label{fig:novellaw-design}
\end{figure}

\lead{NovelLaw creates structural novelty by recombining four interpretable
equation components.} Targets are composed from range, object-context, velocity, and action response
components. Range is fractional-power, screened-power, smooth-gated, or
rational-saturating; context is identity, mass fraction, mass contrast, or
radius fraction; velocity response is linear, range-modulated, gated, or
mass-weighted; and action is absent, additive, range-modulated, or saturating.
Every primitive component occurs in the 16 benchmark development cases. A
frozen combinatorial split then reserves selected range--context and
velocity--action pairings. The IID split reuses four development motifs with
new coefficients and observations; the compositional split contains 12 motifs
with one held-out pairing; and the stress split contains four motifs with two
held-out pairings. Three independently sampled cases per test motif produce 60
final cases. Full compositional and stress motif signatures are disjoint from
development. Figure~\ref{fig:novellaw-design} summarizes the construction and
the separation of discovery evidence from evaluation.

\begin{table}[!htbp]
\caption{\lead{Closed-loop revision improves functional recovery and shifted
prediction over passive baselines.} The full controlled comparison supplements
Table~\ref{tab:rq1}A. NovelLaw reports median NMSE over 60 cases; robot errors
are mean $\pm$ SD over five episode-disjoint partitions. Rejected runs are
included. Bold marks the best result in each column, including ties.}
\label{tab:rq1-full}
\centering
\small
\setlength{\tabcolsep}{2pt}
\renewcommand{\arraystretch}{0.95}
\begin{tabular}{@{}lrrrrrr@{}}
\toprule
& \multicolumn{2}{c}{NovelLaw (60 cases)}
& \multicolumn{2}{c}{PullCubeTool (5 runs)}
& \multicolumn{2}{c}{PushT (5 runs)} \\
\cmidrule(lr){2-3}\cmidrule(lr){4-5}\cmidrule(lr){6-7}
Method & \shortstack{Functional\\recovery (\%, $\uparrow$)}
& \shortstack{Shift\\NMSE ($\downarrow$)}
& \shortstack{Admitted\\($\uparrow$)}
& \shortstack{Shift\\NMSE ($\downarrow$)}
& \shortstack{Admitted\\($\uparrow$)}
& \shortstack{Shift\\NMSE ($\downarrow$)} \\
\multicolumn{7}{@{}l@{}}{\makebox[\linewidth][l]{\textit{Baselines}\enspace\hrulefill}} \\
Fixed/simple model & 0.0 & $5.94\!\times\!10^{-1}$ & 0/5 & $5.773\pm0.287$ & 0/5 & $2.762\pm1.098$ \\
Polynomial SINDy & 28.3 & $1.16\!\times\!10^{-1}$ & 0/5 & $16.573\pm10.539$ & 0/5 & $18.281\pm18.708$ \\
Matched MLP & 0.0 & $4.20\!\times\!10^{-1}$ & 0/5 & $5.570\pm0.407$ & 1/5 & $1.815\pm1.609$ \\
\multicolumn{7}{@{}l@{}}{\makebox[\linewidth][l]{\textit{Variants}\enspace\hrulefill}} \\
One-shot agent & 43.3 & $6.54\!\times\!10^{-2}$ & 0/5 & $4.816\pm1.973$ & 3/5 & $0.842\pm0.644$ \\
Revision-only agent & 75.0 & $1.50\!\times\!10^{-2}$ & 0/5 & $1.296\pm0.024$ & 3/5 & $1.292\pm1.638$ \\
Random-query agent & \textbf{85.0} & $9.15\!\times\!10^{-3}$ & \textbf{5/5} & $0.194\pm0.138$ & \textbf{4/5} & $0.673\pm0.473$ \\
\rowcolor{alderrow}
\textbf{\method} & \textbf{85.0} & $\mathbf{5.58\!\times\!10^{-3}}$ & \textbf{5/5} & $\mathbf{0.161\pm0.052}$ & \textbf{4/5} & $\mathbf{0.420\pm0.185}$ \\
\bottomrule
\end{tabular}
\par\smallskip
{\footnotesize\raggedright
Admitted counts use shared predictive checks, not control successes;
formula proposals also require structural validity. The first three baselines use
initial public evidence. ``Fixed'' denotes the initial NovelLaw library and a
passive linear robot model.\par}

\end{table}

\lead{The frozen split reserves specific cross-component pairings rather than
unseen primitive operators.} The reserved range--context pairs are power--mass-contrast,
screened--radius-fraction, gated--mass-fraction, and rational--identity. The
reserved velocity--action pairs are linear--range-modulated,
range-modulated--saturating, gated--absent, and mass-weighted--additive.

\begin{table}[h]
\caption{\lead{NovelLaw tests new compositions of familiar primitives.}
Each test structure has three
independently sampled parameter/data instances.}
\label{tab:novellaw-splits}
\centering
\small
\begin{tabular}{lrrp{7.2cm}}
\toprule
Split & Structures & Cases & Structural role \\
\midrule
Development & 8 & 16 & Covers every primitive component; used only for protocol and prompt development. \\
IID & 4 & 12 & Reuses a development motif with unseen coefficients and observations. \\
Compositional & 12 & 36 & Contains exactly one reserved component pairing. \\
Stress & 4 & 12 & Contains two simultaneous reserved component pairings. \\
\bottomrule
\end{tabular}
\end{table}

\lead{NovelLaw separates narrow public evidence from wider protected and
functional evaluation domains.} Public fit data cover a narrow domain, with
independent public validation data for each case. Admission widens
interpolation; shift changes distance, mass, radius,
velocity, and action ranges. Gaussian noise has scale 0.5\% of the fit target
standard deviation. A noiseless 4,096-point test set evaluates functional
equivalence. Generator draws are retained only after frozen finiteness,
variance, component-identifiability, complexity, oracle-refit, and
initial-library-exclusion checks. Method outputs are never used for selection.

\lead{All NovelLaw baselines share public evidence but differ in how they obtain
equation structure.} The fixed library has six structures. Polynomial SINDy uses a degree-three
library with its threshold selected on public validation data. The MLP has two
hidden layers and uses public validation data for early stopping. One-shot and closed-loop methods
share the same first proposal; only the latter receives protected rejection
feedback and target-blind selected observations. The oracle receives the target
structure but must refit its coefficients. A public release contains
fit/validation evidence and commitments but excludes targets, protected
outcomes, functional-test rows, and unqueried interactive outcomes.

\lead{The benchmark checks confirm data separation and method-independent
case selection.} All 76 generated cases are rechecked against the predefined
quality criteria, with no overlapping inputs across data splits.
Seventy-one cases pass on their first generator draw; the remaining five
require at most three retries.
The median oracle functional-test NMSE is below $6{\times}10^{-7}$, whereas the median
best initial-library NMSE is 0.274. Case selection never reads \method outputs.
The public release includes public data for all 76 cases but excludes
harness-held and evaluator-only data.

\subsubsection{Agent Variants and Failure Analysis}
\label{app:rq1-auxiliary}

\lead{Auxiliary agent variants measure discovery-loop sensitivity but do not isolate
the query rule.} Revision-only, random-query, and \method use separate sandboxes and independent
model calls with the same proposal and interaction budgets. One-shot is instead
the stored first proposal from the paired \method run. Table~\ref{tab:rq1-agent-aux}
therefore measures end-to-end sensitivity to the discovery loop, while
RQ2 supplies identical candidate sets to isolate acquisition strategy.

\begin{table}[h]
\caption{\lead{Revised proposals recover more equations than one-shot proposals.}
RQ1 values are functional recovery (\%);
queries are medians over all 60 cases. Independent model calls make the last
three rows an auxiliary sensitivity analysis, not a controlled selector
ablation.}
\label{tab:rq1-agent-aux}
\centering
\small
\begin{tabular}{lrrrrrr}
\toprule
Method & IID & Comp. & Stress & All & Shift NMSE & Queries \\
\midrule
One-shot & 41.7 & 50.0 & 25.0 & 43.3 & 0.0654 & 0 \\
Revision-only & 58.3 & 86.1 & 58.3 & 75.0 & 0.0150 & 0 \\
Random-query & 91.7 & 91.7 & 58.3 & 85.0 & 0.00915 & 24 \\
\method & 91.7 & 91.7 & 58.3 & 85.0 & 0.00558 & 24 \\
\bottomrule
\end{tabular}
\end{table}

\begin{table}[h]
\caption{\lead{Paired tests support gains over one-shot and revision-only
discovery.} RQ1 shift-error ratios below one favor closed-loop \method.}
\centering
\small
\begin{tabular}{lrrr}
\toprule
Comparator & Median shift ratio & Wilcoxon $p$ & Recovery wins/losses \\
\midrule
One-shot agent & 0.0782 & $5.18{\times}10^{-9}$ & 25/0 \\
Revision-only & 0.146 & $2.66{\times}10^{-7}$ & 7/1 \\
Random-query & 0.901 & 0.203 & 3/3 \\
Fixed library & 0.00782 & $1.63{\times}10^{-11}$ & 51/0 \\
Polynomial SINDy & 0.0249 & $2.32{\times}10^{-10}$ & 34/0 \\
MLP & 0.0130 & $1.80{\times}10^{-11}$ & 51/0 \\
\bottomrule
\end{tabular}
\end{table}

\lead{Post-hoc failures concentrate in deeper compositions and one repeated
gated-range recipe.} Recovery is 15/15 for short targets, 21/24 for medium targets, and 15/21 for
deep targets. All three instances of the held-out gated-range plus mass-context
recipe fail, and the analogous dual-composition gated stress recipe also fails.
This pattern is consistent with the composition-depth boundary in the main
result. It is descriptive rather than causal because complexity tier and split
are not independently randomized.

\subsubsection{Robot Interaction Protocols}
\label{app:hard-robot-discovery}

\begin{figure}[!t]
\centering
\includegraphics[width=\linewidth]{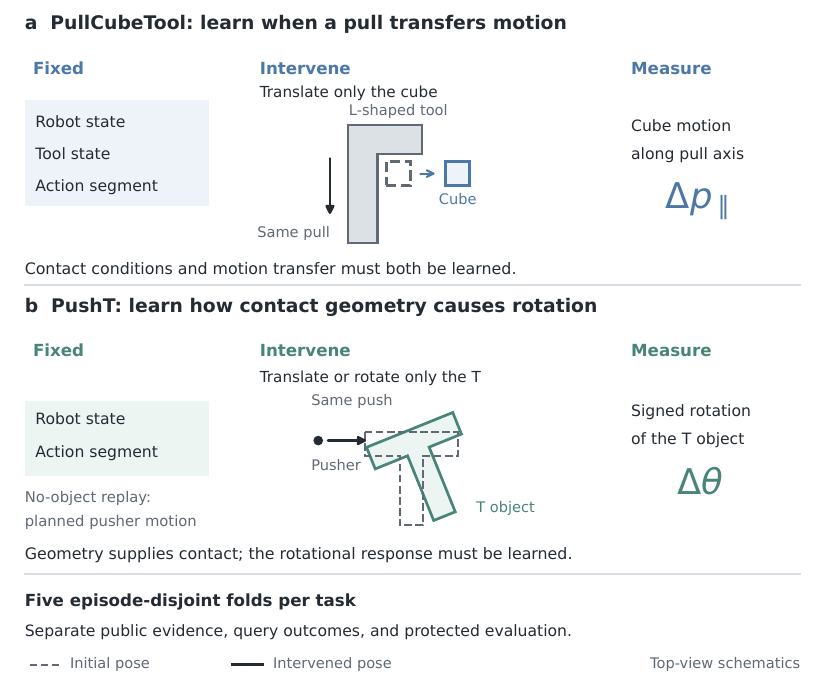}
\caption{\lead{Matched-action interventions probe contact transfer and rotational
response.} (a) PullCubeTool changes only the cube position and measures motion
along the pull axis. (b) PushT changes only the T pose and measures signed
rotation; geometry supplies the contact input, while a no-object replay supplies
planned pusher motion. Both retain the initial robot state and recorded action
segment. Outlines show alternative initial poses, not measured trajectories.}
\label{fig:rq1-robot-design}
\end{figure}

\lead{The hard robot tracks use public environments and official demonstrations, but
the intervention protocols and evidence partitions are constructed by us.}
For PullCubeTool, the experiment harness restores a complete Panda--tool--cube state
immediately before cube motion, translates only the cube in the L-shaped
tool's frame, and replays the identical recorded action segment. Inputs contain
the initial relative pose, local tool and cube velocities, relative axis, and
commanded tool displacement; the target is cube displacement along the pull
axis. Across the five formal folds, 340 source episodes are assigned without
overlap to public, query, admission, and shifted roles. Public examples mostly
show successful transfer, whereas harness-held interventions include full
contact, partial contact, and misses at unseen offsets.
Figure~\ref{fig:rq1-robot-design} contrasts the two robot intervention protocols.

\lead{A PullCubeTool query reveals an incorrect contact gate and drives equation revision.}
Figure~\ref{fig:rq1-example} illustrates a PullCubeTool run.
The lateral coordinate $y$ is measured from the tool-handle centreline to the
cube centre. Transferred fraction divides cube displacement by commanded pull
remaining after closing the estimated initial gap; plotted curves include
fitted transfer gains. Initial evidence lies at 5.8--7.1\,cm. All eight
queries are shown, some overlapping, and the gray dashed line is a no-gate
reference. Near 10\,cm, the initial zero-transfer prediction conflicts with
near-unit observed transfer. Revision lowers protected-admission NMSE from
2.26 to 0.103; shifted-test NMSE is 0.138. The initial 8.08\,cm zero-response
endpoint and revised 11.76\,cm transition midpoint have different definitions
and are not directly comparable thresholds. The lower panels are separate
top-view schematics, not trajectory replays; the displayed formulas show only
the lateral gates, not the complete prediction equations.

\lead{PushT interventions vary contact geometry while replaying the same action to reveal rotational response.}
For PushT, the experiment harness restores an official ManiSkill Panda state near
contact, translates or rotates only the non-convex T, and replays the same
action. A paired no-object replay supplies intended TCP displacement without
revealing the contact response. Inputs contain pusher pose in the initial T
frame, the paired TCP displacement, and a swept-contact predicate computed from
the known T geometry; the target is signed T rotation. The five folds use 362
disjoint source episodes. Public evidence emphasizes weak contacts, while
queries and protected splits vary approach direction, T orientation, and
off-center lever arm.

\lead{The passive difficulty checks verify both ambiguity and representability
before any proposal-agent result is considered.} Table~\ref{tab:hard-robot-passive}
shows that a linear model, polynomial SINDy, and matched MLP extrapolate
poorly from public evidence. A harness-supplied equation structure also remains
ambiguous when its coefficients are fitted only on public rows, but becomes
accurate after receiving the full query pool. These supplied structures are
benchmark diagnostics and are never presented to the proposal agent.

\begin{table}[h]
\caption{\lead{Query evidence improves prediction even when structure is
supplied.} Hard-track entries are mean shifted NMSE over five folds; lower is
better. ``All queries'' uses every initially unrevealed query outcome and is not an
\method discovery result.}
\label{tab:hard-robot-passive}
\centering
\small
\begin{tabular}{lrr}
\toprule
Model and evidence & PullCubeTool & PushT \\
\midrule
Linear, public & 5.773 & 2.762 \\
Polynomial SINDy, public & 16.573 & 18.281 \\
MLP, public & 5.570 & 1.815 \\
Supplied structure, public & 4.579 & 0.919 \\
Supplied structure, all queries & \textbf{0.212} & \textbf{0.588} \\
\bottomrule
\end{tabular}
\end{table}

\lead{Each formal agent grid contains the same four conditions and five folds.}
One-shot receives only the initial public evidence; revision-only receives the
same rejection ledger but no new outcome; random-query and \method may reveal
at most the same number of intervention outcomes. All use the same typed
grammar, numerical fitter, proposal budget, admission thresholds, and protected
splits. All 40 runs request \texttt{codex-default} with high reasoning effort,
and each manifest records its CLI version and requested model setting.
Table~\ref{tab:hard-robot-agent} retains rejected runs in every mean.

\begin{table*}[!htbp]
\caption{\lead{Interactive revision improves model admission on both robot
tracks.} Values are mean $\pm$
standard deviation over five folds. MAE is in centimeters for PullCubeTool
and radians for PushT.}
\label{tab:hard-robot-agent}
\centering
\small
\setlength{\tabcolsep}{3pt}
\begin{tabular}{@{}lrrr|rrr@{}}
\toprule
& \multicolumn{3}{c}{PullCubeTool} & \multicolumn{3}{c}{PushT} \\
\cmidrule(lr){2-4}\cmidrule(lr){5-7}
Method & Admitted & Shift NMSE & Shift MAE & Admitted & Shift NMSE & Shift MAE \\
\midrule
One-shot & 0/5 & $4.816\pm1.973$ & $1.081\pm0.368$ & 3/5 & $0.842\pm0.644$ & $0.086\pm0.032$ \\
Revision-only & 0/5 & $1.296\pm0.024$ & $0.377\pm0.032$ & 3/5 & $1.292\pm1.638$ & $0.106\pm0.072$ \\
Random-query & \textbf{5/5} & $0.194\pm0.138$ & $0.144\pm0.064$ & \textbf{4/5} & $0.673\pm0.473$ & $0.078\pm0.021$ \\
\rowcolor{alderrow}
\textbf{\method} & \textbf{5/5} & $\mathbf{0.161\pm0.052}$ & $\mathbf{0.136\pm0.029}$ & \textbf{4/5} & $\mathbf{0.420\pm0.185}$ & $\mathbf{0.071\pm0.024}$ \\
\bottomrule
\end{tabular}
\end{table*}

\lead{Interactive revision improves model admission and shifted prediction relative to non-interactive baselines.}
Across these two tracks, \method passes 9/10 paired folds, compared with 3/10
for one-shot and revision-only. Against either non-interactive condition it
has six admission wins, four ties, and no losses (two-sided exact $p=0.0312$),
while its shifted NMSE is lower on all ten folds ($p=0.0020$). Random-query
also passes 9/10; \method wins six and loses four shifted-NMSE comparisons
against it ($p=0.7539$). Since random-query and \method use independent proposal
calls, this grid tests the complete discovery loop rather than the
isolated selector. RQ2 supplies that controlled comparison. An independent
audit checks all 40 method--fold runs and 76 proposal rounds: every agent
executes inside the declared Bubblewrap namespace, no protected or
harness-held file appears in its repository, and no failed proposal is
removed.

\FloatBarrier

\subsubsection{Equation Discovery from Shared NovelLaw Observations}
\label{app:novellaw-shared-observations}

\lead{Sharing public observations acquired by \method enables a same-data
comparison of equation discovery.} On all 60 NovelLaw cases, we provide
SINDy, PySR, LLM-SR, and SR-Scientist with the final public fitting observations
acquired by \method and use the same public development set for model
selection. The baselines issue no additional queries and receive neither
\method's equation proposals, fitted parameters, nor admission feedback.
\method uses its previously saved and frozen final models as the reference.

\lead{Each method retains its own search procedure, with computational budgets
and expression settings recorded explicitly.} SINDy uses a degree-three
polynomial library. PySR is evaluated with a base operator configuration and
with a configuration that adds real powers and increases the search budget:
the former allows at most 100 iterations and $10^5$ expression evaluations,
and the latter allows 1,000 iterations and $10^6$ evaluations. Both are limited
to 31 nodes and depth 8. LLM-SR generates 3 or 12 proposals; the 12-proposal
configuration also expands program capacity, so the difference is not due
solely to proposal count. SR-Scientist allows at most 3 or 12 model turns,
with at most two tool calls per turn. The added LLM baselines use a common
\texttt{gpt-5.6-sol} backend.

\lead{After all models are fixed, common new test points measure prediction
error and functional recovery.} Each case is evaluated on 4,096 new points
in the Shift domain and 4,096 in the broader functional-test domain. These
points are not used for fitting, model selection, or admission. Functional
recovery requires variance-normalized NMSE at most 0.02 in the corresponding
test domain and does not imply an exact symbolic match. All cases and failed
predictions remain in the denominator.

\begin{table}[!htbp]
\caption{\lead{Equation discovery from shared observations on 60 NovelLaw cases.}}
\label{tab:novellaw-shared-observations}
\centering
\small
\setlength{\tabcolsep}{4pt}
\begin{tabularx}{0.93\linewidth}{>{\raggedright\arraybackslash}Xrrrr}
\toprule
& \multicolumn{2}{c}{Shift}
& \multicolumn{2}{c}{Broader functional domain} \\
\cmidrule(lr){2-3}\cmidrule(lr){4-5}
Method and configuration & Recovery $\uparrow$ & Median NMSE $\downarrow$
& Recovery $\uparrow$ & Median NMSE $\downarrow$ \\
\midrule
SINDy & 29/60 & 0.02279 & 30/60 & 0.02005 \\
PySR: base configuration & 1/60 & 0.18997 & 3/60 & 0.19473 \\
PySR: real powers, larger search budget & 28/60 & 0.02394 & 30/60 & 0.01928 \\
LLM-SR: 3 proposals & 9/60 & 0.22754 & 12/60 & 0.11776 \\
LLM-SR: 12 proposals, expanded program capacity & 21/60 & 0.03486 & 26/60 & 0.03022 \\
SR-Scientist: at most 3 turns & 10/60 & 0.15126 & 10/60 & 0.11702 \\
SR-Scientist: at most 12 turns & 36/60 & 0.01346 & 39/60 & 0.00762 \\
\rowcolor{alderrow}
\method: frozen reference & \textbf{50/60} & \textbf{0.00630}
& \textbf{51/60} & \textbf{0.00515} \\
\bottomrule
\end{tabularx}
\end{table}

\lead{With shared observations, the frozen \method reference achieves the
highest functional recovery and lowest median NMSE in both test domains.}
Expanded operator sets, search budgets, and program capacity also
substantially improve some baselines. These configurations are therefore
reported separately rather than attributing their differences to a single
budget factor.

\lead{Comparing each method under two observation conditions examines how
additional evidence affects discovery.} On all 60 NovelLaw cases, we use
either the initial public fitting observations alone or those observations
augmented with public observations acquired by \method. Both conditions use
the same public development set, with each method's search configuration
held fixed and no additional queries issued. Once the models are fixed, we
apply the common final evaluation described above. The table reports PySR
with real powers and a larger search budget, LLM-SR with 12 proposals and
expanded program capacity, and SR-Scientist with at most 12 turns.

\begin{table}[!htbp]
\caption{\lead{Additional observations increase Shift recovery for all four
discovery methods.}}
\label{tab:novellaw-additional-observations}
\centering
\small
\setlength{\tabcolsep}{4pt}
\begin{tabularx}{0.93\linewidth}{>{\raggedright\arraybackslash}Xrr}
\toprule
Method and configuration & \shortstack{Shift recovery:\\passive only}
& \shortstack{Shift recovery:\\with \method-acquired\\observations} \\
\midrule
SINDy & 13/60 & 29/60 \\
PySR: real powers, larger search budget & 15/60 & 28/60 \\
LLM-SR: 12 proposals, expanded program capacity & 8/60 & 21/60 \\
SR-Scientist: at most 12 turns & 21/60 & 36/60 \\
\bottomrule
\end{tabularx}
\end{table}

\lead{All four methods recover more Shift cases when \method-acquired
observations are included.} Functional recovery still requires Shift NMSE
at most 0.02, with all cases retained in the denominator. These results show
that observations acquired by \method can also improve recovery by other
discovery methods.

\FloatBarrier

\subsubsection{Paired Query Comparisons in Complete NovelLaw Discovery}
\label{app:novellaw-paired-queries}

\lead{The paired experiment and the original functional-recovery results use
different run batches and final evaluation domains.} The original results
evaluate final models from the original \method runs in the functional-recovery
test domain (semantic). Here, we retain the same target set but run paired
continuations from initial equations and fitted parameters frozen in the new
experiment, and evaluate on newly sampled independent Shift test points.
Admission in this protocol does not use the old Shift set. This is therefore
neither a simple repeat evaluation of the original final models nor a
like-for-like replication of the original functional-recovery results; the
two sets of results are reported separately and do not replace one another.

\lead{Sharing initial equations and fitted parameters gives both query
strategies the same starting point for continued discovery.} We use all 60
NovelLaw cases with three numbered repetitions per case, yielding 180 paired
starting points. Each pair of ARS and random-query branches reuses the same
frozen initial equation and fitted parameters, with the same expression space,
admission rules, and observation budget, and then separately performs querying,
fitting, and equation revision. Each run allows at most three outer equation
proposal rounds. If the initial proposal fails admission, each batch adds 24
observation points, with at most two batches and 48 additional points in total.
Runs stop on admission or budget exhaustion.

\lead{Final functional recovery is evaluated on common test points excluded
from discovery.} After all final models are frozen, prediction error is
computed on 4,096 new Shift test points per case. Functional recovery requires
variance-normalized NMSE at most 0.02 and does not imply an exact symbolic
match. Test points are not used for fitting, admission, or model selection,
and all runs and failures remain in the statistics.

\begin{table}[H]
\caption{\lead{ARS uses fewer observations with similar functional recovery
in this paired batch.}}
\label{tab:novellaw-paired-queries}
\centering
\small
\begin{tabularx}{0.93\linewidth}{l*{3}{>{\raggedleft\arraybackslash}X}}
\toprule
Query strategy & \shortstack{Independent Shift\\functional recovery $\uparrow$}
& \shortstack{Median Shift\\NMSE $\downarrow$}
& \shortstack{Mean additional\\observation points $\downarrow$} \\
\midrule
ARS & 97/180 & 0.01753 & 3.33 \\
Random querying & 95/180 & 0.01708 & 3.73 \\
\bottomrule
\end{tabularx}
\end{table}

\lead{Mean observation counts include all runs admitted initially without
further queries.} Of the 180 shared starting points, 159 pass initial admission
and the remaining 21 proceed to further querying and equation revision.
Passing initial admission does not imply functional recovery on the independent
Shift test. Observation counts include all zero-query runs and are not
conditioned on final recovery success.

\lead{In this paired batch, ARS uses fewer observations to achieve similar
functional recovery as random querying.} Compared with random querying,
ARS reduces the mean number of additional observation points by 10.7\%.
This result reflects an observation-efficiency gain in this batch of complete
discovery runs rather than an improvement in recovery rate.

\FloatBarrier

\subsection{ARS Reduces Mean Identification Cost}
\label{app:intervention-selection}

\subsubsection{Controlled Model Identification}

\lead{The intervention protocol isolates information acquisition with matched
candidate models, query pools, and budgets.} The candidate models vary contact gating, mass--action scaling, action
response, and coupled contact response. The safe query pool spans distance,
radial speed, action, and mass ratio. It contains geometrically novel points
that are irrelevant to the candidate disagreement so that geometric coverage
is nontrivial. Candidate coefficients are refitted after each observation.
Passive sampling stays near the initial observation regime. Correct
identification must be sustained at high confidence. The identification-cost
score is the first query count satisfying that criterion for a successful
case; an unresolved case receives nine, one above the eight-query budget.
All means, medians, and paired reductions use this same score, including
failures. They measure penalized identification cost rather than the total
number of queries executed. The oracle uses the target model index to
choose a query that separates it from alternatives; this greedy reference is
not a theoretical lower bound on identification cost.

\lead{Identification curves measure retrospectively sustained confidence, not online stopping times.}
Figure~\ref{fig:rq2} plots the empirical cumulative distribution of sustained
identification times. Every curve uses the same 182 initially ambiguous
family--seed cases; unresolved trials remain in the denominator at every
query count. A time is the first correct, high-confidence round after which
every remaining round is also correct at high confidence. All trials execute
the full eight-query budget, so these retrospective times do not describe an
online early-stopping policy. The curves are reconstructed from the frozen
round-by-round confidence records without interpolation. Table~\ref{tab:rq2}
retains the complete aggregate comparison, and
Table~\ref{tab:rq2-paired} reports its paired uncertainty estimates.

\begin{table}[!htbp]
\caption{\lead{ARS identifies all targets with the lowest mean penalized
cost.} RQ2 uses 182 initially ambiguous cases, the same candidate models,
and an eight-query budget. Bold marks the best non-oracle results, including
ties.}
\label{tab:rq2}
\centering
\small
\setlength{\tabcolsep}{5pt}
\renewcommand{\arraystretch}{1.08}
\begin{tabular}{lrrr}
\toprule
Acquisition & \shortstack{Identified\\($\uparrow$)}
& \shortstack{Mean cost$^\dagger$\\($\downarrow$)}
& \shortstack{Median cost$^\dagger$\\($\downarrow$)} \\
\midrule
Passive sampling & 4.4\% & 8.85 & 9 \\
Random selection & 94.0\% & 3.53 & 3 \\
Geometric coverage & \textbf{100.0\%} & 1.91 & \textbf{1} \\
Parameter uncertainty & 99.5\% & 1.35 & \textbf{1} \\
\rowcolor{alderrow}
\textbf{\method (ARS)} & \textbf{100.0\%} & \textbf{1.16} & \textbf{1} \\
\midrule
Oracle reference & 100.0\% & 1.18 & 1 \\
\bottomrule
\end{tabular}
\par\smallskip
{\footnotesize\raggedright
$\dagger$ \lead{Identification cost includes a penalty for unresolved cases.}
Cost measures retrospective sustained identification, not total executed
queries; unresolved cases receive nine (budget plus one). The oracle is a
reference, not an optimal-cost bound. Paired results appear in
Table~\ref{tab:rq2-paired}.\par}
\end{table}

\begin{table}[h]
\caption{\lead{ARS reduces mean identification cost relative to each listed
comparator.} Paired RQ2 reductions are positive when they favor ARS.}
\label{tab:rq2-paired}
\centering
\small
\begin{tabular}{lrr}
\toprule
Comparator & Mean cost reduction [95\% CI] & Win/tie/loss \\
\midrule
Passive & 7.69 [7.55, 7.80] & 182/0/0 \\
Random & 2.36 [2.05, 2.68] & 151/29/2 \\
Coverage & 0.74 [0.56, 0.94] & 63/118/1 \\
Parameter uncertainty & 0.18 [0.08, 0.32] & 18/161/3 \\
\bottomrule
\end{tabular}
\end{table}

\subsubsection{Online Stopping and Query Pool Comparison}
\label{app:online-stopping}

\lead{Online stopping based on current confidence tests actual query
efficiency.} We compare ARS and random selection with fixed candidate
equations. Both methods share the initial observations, candidate equations,
and a budget of at most eight additional queries. They stop when the largest
posterior probability among the candidate models reaches 0.9; otherwise, they
continue until the query budget is exhausted. Stopping decisions use only the
observations revealed so far.

\lead{Three query pools test sensitivity to the distribution of candidate
interventions.} We use the original mixed pool, a uniformly sampled pool, and
a Latin-hypercube pool. Each pool is evaluated on four mechanism families with
50 seeds per family, giving 200 cases. All cases are retained, including those
that can stop after the initial observations; incorrect high-confidence stops
count as identification failures. Mean query counts are actual additional
queries executed over all cases, without a failure penalty or the shared
initial observations. Each query comprises two ten-step trajectories.

\begin{table}[H]
\caption{\lead{ARS achieves higher correct-identification rates with fewer
mean queries in all three pools.}}
\label{tab:online-stopping-pools}
\centering
\small
\begin{tabularx}{0.93\linewidth}{l*{4}{>{\raggedleft\arraybackslash}X}}
\toprule
Query pool & \shortstack{ARS\\identified $\uparrow$}
& \shortstack{Random\\identified $\uparrow$}
& \shortstack{ARS\\mean queries $\downarrow$}
& \shortstack{Random\\mean queries $\downarrow$} \\
\midrule
Original mixed & 194/200 & 180/200 & 1.085 & 3.080 \\
Uniform & 195/200 & 189/200 & 1.085 & 1.735 \\
Latin-hypercube & 194/200 & 186/200 & 1.065 & 1.830 \\
\bottomrule
\end{tabularx}
\end{table}

\lead{The online query-efficiency gains are not limited to the original mixed
pool.} With uniformly sampled and Latin-hypercube query pools, ARS also uses
fewer queries than random selection and correctly identifies more cases.
These results support its online identification efficiency in the tested
fixed-candidate equation setting.

\subsubsection{RoboCasa Selector Comparison}
\label{app:robocasa-selectors}

\lead{The RoboCasa selector comparison varies acquisition while holding
candidate equation structures fixed.} It uses the worlds specified in
Appendix~\ref{app:robocasa-protocol}. Table~\ref{tab:robocasa-selectors} shows that querying is
essential for Coulomb and mixed worlds, but does not establish universal
superiority of prediction disagreement over coverage or uncertainty. All active
selectors eventually identify every Door world, while random selection uses a
median of three queries compared with 1.5 for the structured selectors.

\begin{table*}[!htbp]
\caption{\lead{Structured queries identify Door models sooner than random
queries.} RoboCasa candidates are fixed. Queries are censored at seven after
exhausting the six-query budget. ``Support'' reports exact recovery of the
optional equation terms after coefficient thresholding.}
\label{tab:robocasa-selectors}
\centering
\small
\setlength{\tabcolsep}{6pt}
\begin{tabular}{llrrrr}
\toprule
Task & Selector & Identified & Support & Median queries & Safety \\
\midrule
Drawer & Passive & 50.0\% & 50.0\% & 3.5 & 0 \\
 & Random & 100.0\% & 97.5\% & 0.5 & 0 \\
 & Coverage & 100.0\% & 70.0\% & 0.5 & 0 \\
 & Uncertainty & 100.0\% & \textbf{100.0\%} & 0.5 & 0 \\
 & \textbf{ALDER} & 100.0\% & 97.5\% & 0.5 & 0 \\
\addlinespace
Door & Passive & 25.0\% & 25.0\% & 7.0 & 0 \\
 & Random & 100.0\% & 100.0\% & 3.0 & 0 \\
 & Coverage & 100.0\% & 100.0\% & \textbf{1.5} & 0 \\
 & Uncertainty & 100.0\% & 100.0\% & \textbf{1.5} & 0 \\
 & \textbf{ALDER} & 100.0\% & 100.0\% & \textbf{1.5} & 0 \\
\bottomrule
\end{tabular}

\end{table*}

\subsubsection{Model Admission at Query Checkpoints}
\label{app:query-budget-prefix}

\lead{Query checkpoints show how additional interactions accumulate admitted
models.} We retrospectively analyze the 80 frozen runs of ARS and random
selection across all 40 Drawer worlds, comparing outcomes after zero to six
additional queries. Both strategies share the initial proposal, fitted
equation, and initial admission decision. All checkpoints come from these
original runs.

\lead{Admission and term-support matching measure model usability and
structural recovery separately.} Cumulative admission counts worlds with a
model that has passed verification by the checkpoint. Admission with correct
term support additionally requires the original evaluator's strict
term-support match. All 40 worlds remain in the denominator.

\begin{table}[H]
\caption{\lead{ARS reaches admission across all Drawer worlds earlier in the
original runs.}}
\label{tab:query-budget-prefix}
\centering
\small
\begin{tabularx}{0.93\linewidth}{r*{4}{>{\raggedleft\arraybackslash}X}}
\toprule
& \multicolumn{2}{c}{Admitted}
& \multicolumn{2}{c}{\shortstack{Admitted with\\correct term support}} \\
\cmidrule(lr){2-3}\cmidrule(lr){4-5}
\shortstack{Additional\\queries} & ARS & Random & ARS & Random \\
\midrule
0 & 20/40 & 20/40 & 14/40 & 14/40 \\
1 & 32/40 & 31/40 & 26/40 & 25/40 \\
2 & 37/40 & 34/40 & 31/40 & 28/40 \\
3 & 40/40 & 38/40 & 34/40 & 31/40 \\
4 & 40/40 & 39/40 & 34/40 & 32/40 \\
5 & 40/40 & 40/40 & 34/40 & 32/40 \\
6 & 40/40 & 40/40 & 34/40 & 32/40 \\
\bottomrule
\end{tabularx}
\end{table}

\lead{ARS reaches admission across all worlds earlier in the original runs.}
ARS admits models in all 40/40 worlds within three additional queries, while
random selection does so within five. Both strategies admit models in 20/40
worlds without additional queries, explaining the low overall query counts.

\lead{Initially rejected worlds allow query strategies to be compared when
additional evidence is needed.} We analyze the 20 worlds in the original
paired Drawer runs whose shared initial proposal failed admission. Each pair
of ARS and random-query branches shares the initial equation, fitted parameters,
and initial admission decision, with the same expression space, fitting
procedure, admission rules, and budget of at most six additional queries.
Each query corresponds to one intervention rollout. After obtaining
observations, both methods separately continue equation revision, parameter
fitting, and admission checks.

\lead{Admission with correct term support requires both passing verification
and recovering the correct equation terms.} Term support follows the original
evaluator's strict matching rule. Mean additional query counts cover the
queries actually executed across all 20 worlds in this subset and are not
conditioned on whether the final term support is correct.

\begin{table}[H]
\caption{\lead{ARS obtains more admitted models with correct term support
using fewer queries on average.}}
\label{tab:drawer-initially-rejected}
\centering
\small
\begin{tabularx}{0.93\linewidth}{l*{2}{>{\raggedleft\arraybackslash}X}}
\toprule
Query strategy & \shortstack{Admitted with\\correct term support $\uparrow$}
& \shortstack{Mean additional\\queries $\downarrow$} \\
\midrule
ARS & 20/20 & 1.55 \\
Random querying & 18/20 & 1.90 \\
\bottomrule
\end{tabularx}
\end{table}

\lead{In this subset requiring further queries, ARS obtains more admitted
models with correct term support and uses fewer queries on average.}
Compared with random querying, ARS reduces the mean number of additional
queries by 18.4\%. These numbers are reported as a descriptive comparison
within this subset.

\lead{Worlds admitted initially explain the low query counts in the overall
study.} The other 20 Drawer worlds pass initial admission and require no
additional queries. Including these zero-query worlds yields a median of
0.5 additional queries over all 40 worlds, while the subset analysis separately
reports discovery when the initial evidence is insufficient.

\subsubsection{Cost and Predictive Safety Screening Ablations}
\label{app:cost-safety-ablations}

\lead{Cost normalization tests whether the selector can reduce intervention
execution steps.} We select two seeds and all four mechanism families in each
of Drawer and Door, giving eight worlds per task. The fixed-candidate
identification controller runs with and without cost normalization, while all
other fitting, admission, and stopping rules remain unchanged. Intervention
cost is the sum of main-action and preliminary-action execution steps. Mean
query counts and execution steps include worlds requiring no additional
queries and exclude shared initialization and reset overhead.

\begin{table}[H]
\caption{\lead{Cost normalization reduces execution steps in the small paired
comparison.}}
\label{tab:cost-normalization-ablation}
\centering
\small
\begin{tabularx}{0.93\linewidth}{ll*{3}{>{\raggedleft\arraybackslash}X}}
\toprule
Task & \shortstack{Cost\\normalization}
& \shortstack{Admitted with\\correct term\\support}
& \shortstack{Mean additional\\queries}
& \shortstack{Mean execution\\steps} \\
\midrule
Drawer & Enabled & 8/8 & 0.50 & 3.0 \\
Drawer & Disabled & 8/8 & 0.50 & 5.0 \\
Door & Enabled & 8/8 & 1.25 & 7.5 \\
Door & Disabled & 8/8 & 1.25 & 9.5 \\
\bottomrule
\end{tabularx}
\end{table}

\lead{Cost normalization reduces execution steps in this small comparison
while preserving query counts and admission outcomes.} Mean execution steps
fall from 5.0 to 3.0 on Drawer and from 9.5 to 7.5 on Door.

\lead{Predictive safety screening ablations test effects on query selection
and identification.} We compare screening-enabled and screening-disabled
branches on 40 fixed-candidate cases comprising four mechanism families and
ten seeds per family. Both branches use the same initial observations,
fitting procedure, and online stopping rule, while retaining the same
numerical bounds and observation-acceptance rules.

\begin{table}[H]
\caption{\lead{Both screening conditions yield the same identification and
violation outcomes.}}
\label{tab:predictive-safety-ablation}
\centering
\small
\begin{tabularx}{0.93\linewidth}{l*{3}{>{\raggedleft\arraybackslash}X}}
\toprule
\shortstack{Predictive safety\\screening}
& \shortstack{Correct confident\\identifications}
& \shortstack{Mean additional\\queries}
& \shortstack{Unsafe\\queries} \\
\midrule
Enabled & 38/40 & 1.125 & 0/45 \\
Disabled & 38/40 & 1.125 & 0/45 \\
\bottomrule
\end{tabularx}
\end{table}

\lead{Predictive screening changes some query choices, but this comparison
shows no improvement in identification or violation rates.} Along the
screening-enabled branch, the filter excludes 44 of 17,273 candidate
evaluations and affects nine of 45 query decisions. Both branches yield the
same correct-identification count, mean query count, and observed number of
unsafe queries.

\FloatBarrier

\subsection{Equation-Based Models Improve Shifted Prediction and Support Control}
\label{app:robocasa-protocol}

\subsubsection{Environments and Evidence Collection}

\lead{The robot study uses physically reached states and matched worlds within each articulation.}
The end-to-end robot study uses RoboCasa 1.0.1 at commit
\texttt{a07e365c} and robosuite 1.5.2 at commit \texttt{5ce6643f}. We instantiate
the prismatic \texttt{OpenDrawer} and revolute \texttt{OpenDoor} tasks with the
same PandaOmron robot, a 20 Hz controller, state observations, and no rendering.
A deterministic common controller reaches and grasps each handle, physically
visits four normalized articulation positions, and saves the resulting MuJoCo
states. Every scored rollout begins from a reached state; direct teleportation
is not used to construct a grasp. Drawer evaluation seeds are 1701--1710 and
Door seeds are 2701--2710. Within an articulation, only the hidden joint
dynamics change across matched worlds.

\begin{figure}[!t]
\centering
\includegraphics[width=\textwidth]{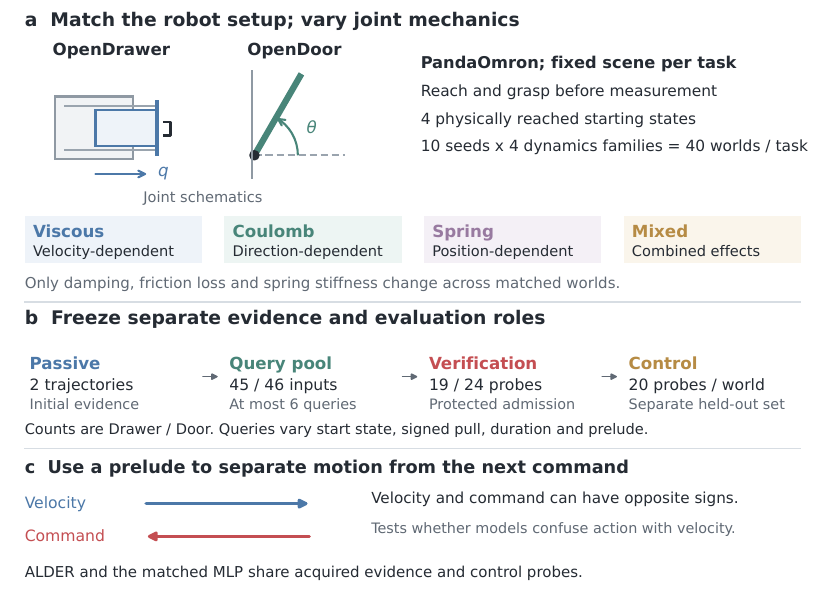}
\caption{\lead{Matched RoboCasa worlds isolate joint mechanics and separate discovery from control evaluation.}
(a) Joint schematics illustrate the two post-grasp tasks and four dynamics
families. (b) Passive, query, admission, and control probes have separate roles;
paired counts denote Drawer and Door. (c) A prelude can make motion oppose the
next command, exposing explanations that substitute action direction for
velocity direction. Arrows illustrate this intervention, not a measured rollout.}
\label{fig:robocasa-design}
\end{figure}

\lead{Both articulations use matched dynamics families and a shared evidence protocol.}
Each articulation contains 40 worlds: ten seeds for viscous damping, Coulomb
friction, linear spring stiffness, and their composition. Every world has two
narrow passive trajectories, a safe query pool, protected admission probes, and
20 disjoint control probes. Drawer provides 45 query candidates and 19
admission probes; Door provides 46 and 24. A nominal-world screen fixes legal
probe IDs before any shifted outcome is evaluated. Queries vary initial
position, signed pull, horizon, and an optional prelude that can make velocity
direction disagree with action direction, testing whether models confuse these
directions when predicting friction. All
methods receive the same cached outcomes and may request at most six queries.
Figure~\ref{fig:robocasa-design} summarizes the matched worlds and probe roles.

\subsubsection{Equation Discovery and Control Evaluation}

\lead{Joint-force equation discovery expands a simple initial program through bounded structural proposals and separate coefficient fitting.}
The initial program has bias and velocity terms. In each round, the proposal
agent may add at most two terms from a declared typed grammar, numerically
compare at most four self-proposed structures, and submit one structure fitted
on public evidence. Candidate primitives include position, friction sign,
smooth sign, speed magnitude, quadratic velocity or position, action
interactions, and an exponential position feature. The complete reference
equation is absent from the sandbox. Harness-owned ridge fitting estimates
coefficients, and held-out normalized error controls admission. The primary
proposal backend is \texttt{gpt-5.6-sol} with frozen reasoning settings. The
same generic joint interface and proposal schema are used for Drawer and Door.

\lead{The original finite-action comparison shares acquired evidence between the MLP and ALDER.}
For control, the passive law is fitted only to passive evidence, whereas the
matched MLP and ALDER law use exactly the passive plus ALDER-selected evidence.
The MLP has two 32-unit tanh layers and fixed regularization. Every predictor
forecasts the same frozen control probes. For each staged state, the planner
forms targets at 20, 40, 60, and 80 percent of the reachable articulation range
and selects the action whose predicted final position is closest. Relative success means
that executed error is within 0.01 normalized position of the best available
candidate. A rejected model refuses all targets in that world and receives zero
relative control success. The parameter oracle knows hidden joint coefficients but not
cached outcomes; the outcome oracle directly ranks those outcomes. Each task
contains 3,840 matched model--target rows across the six control models.

\begin{table}[!t]
\caption{\lead{Equation-based models improve prediction under the original
finite-action evaluation.} Values are mean $\pm$ SD over ten seeds, each
averaging four dynamics families. Relative success means executed target
error within 0.01 of the best finite candidate's error; it is not the absolute
target-reaching success in Table~\ref{tab:robocasa-control}. Each task has 640
targets per model. Bold marks the best non-oracle results, including ties.}
\label{tab:robocasa-discrete-control}
\centering
\small
\setlength{\tabcolsep}{3pt}
\renewcommand{\arraystretch}{1.08}
\begin{tabular}{@{}lrrrr@{}}
\toprule
& \multicolumn{2}{c}{OpenDrawer} & \multicolumn{2}{c}{OpenDoor} \\
\cmidrule(lr){2-3}\cmidrule(lr){4-5}
Model & \shortstack{OOD prediction\\RMSE ($\downarrow$)}
& \shortstack{Relative success\\(\%, $\uparrow$)}
& \shortstack{OOD prediction\\RMSE ($\downarrow$)}
& \shortstack{Relative success\\(\%, $\uparrow$)} \\
\midrule
Nominal & $0.0068\pm0.0007$ & $96.7\pm1.4$ & $0.0072\pm0.0008$ & $95.9\pm2.0$ \\
Passive law & $0.0075\pm0.0010$ & $94.2\pm2.1$ & $0.0074\pm0.0015$ & $96.4\pm2.4$ \\
Matched MLP & $0.0037\pm0.0008$ & $99.5\pm1.1$ & $0.0031\pm0.0007$ & $\mathbf{100.0\pm0.0}$ \\
\rowcolor{alderrow}
\textbf{ALDER law} & $\mathbf{0.0009\pm0.0001}$ & $\mathbf{100.0\pm0.0}$ & $\mathbf{0.0009\pm0.0001}$ & $\mathbf{100.0\pm0.0}$ \\
\midrule
Parameter oracle & $0.0009\pm0.0001$ & $100.0\pm0.0$ & $0.0009\pm0.0001$ & $100.0\pm0.0$ \\
Outcome oracle & $0.0000\pm0.0000$ & $100.0\pm0.0$ & $0.0000\pm0.0000$ & $100.0\pm0.0$ \\
\bottomrule
\end{tabular}

\end{table}

\lead{The original RoboCasa uncertainty estimates use paired seeds, with failed proposals and prediction coverage explicitly accounted for.}
All confirmatory summaries use the ten seeds within each articulation as the
paired bootstrap unit with 10,000 resamples. Admission and term-support recovery
retain every failed proposal in the denominator. Continuous prediction metrics
are conditional on both compared methods producing finite predictions and are
reported with coverage.

\lead{Numerical inverse control tests the learned models against absolute
target-reaching criteria.} The extension retains all 80 worlds and 1,280
targets, with discovered equations, admission decisions, and training evidence
unchanged. All model-based methods use the same bounded Brent solver, action
ranges, and horizon order (11, 17, then 7 steps), with at most 48 forward-model
evaluations per target. A root must have predicted target error at most
$10^{-4}$ and satisfy predicted safety constraints; otherwise, the solver
returns the closest safe evaluated action or refuses execution. Actual
outcomes are evaluated only after action selection. Table~\ref{tab:robocasa-control}
retains prediction RMSE on the original fixed OOD probes and adds mean
absolute executed endpoint-to-target error and success under numerical
inversion. Success requires actual target error at most 0.01 and safe
execution; refusals count as failures. Means and sample SDs use ten seeds per
task, each averaging four families and 16 targets per family. These success
rates are not directly comparable to the original relative-success metric.

\lead{The known-parameter model remains a reference under the same inverse
solver.} It achieves $100.0\pm0.0\%$ success on both tasks, with control MAE
$(0.74\pm0.10)\times10^{-3}$ on Drawer and
$(0.50\pm0.06)\times10^{-3}$ on Door. The outcome oracle remains confined to
the finite-action comparison in Table~\ref{tab:robocasa-discrete-control}; it
is not an optimum over the continuous action range.

\lead{The complete agent loop transfers without changing its interface between the
two articulations.} Across the 40 worlds in each task, every submitted ALDER model
passes protected admission. Strict term-support recovery is 85.0\% on Drawer
and 95.0\% on Door, and both tasks use a median of 0.5 new queries. The geometric
mean protected NMSE is $5.65\times10^{-8}$ and $1.92\times10^{-7}$,
respectively. These rates evaluate open structural proposal;
Appendix~\ref{app:robocasa-selectors} separately isolates acquisition using
fixed candidate equation structures.

\subsubsection{Dynamics-Family Results}

\begin{figure}[!t]
\centering
\includegraphics[width=\textwidth]{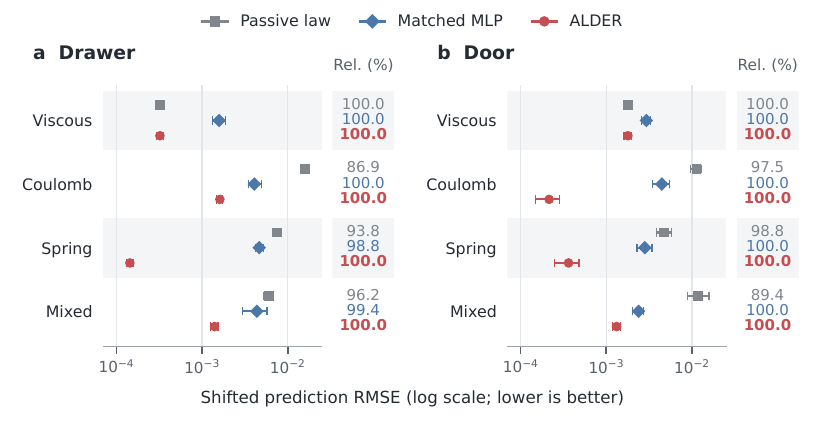}
\caption{\lead{\method predicts more accurately than the MLP in every tested
robot family.} Points show mean shifted prediction RMSE on a shared logarithmic
scale; whiskers are 95\% percentile-bootstrap intervals over ten matched seeds
(10,000 resamples). The adjacent columns retain relative finite-action
success for each method in the same row order.}
\label{fig:robocasa-control-family}
\end{figure}

\lead{ALDER reduces prediction error in every tested task--family cell,
while relative finite-action success is largely saturated.}
Figure~\ref{fig:robocasa-control-family} reports every registered family for
both articulations under the original finite-action evaluation. The updated frozen reproduction gives ALDER lower
prediction RMSE than the matched MLP in all eight task--family cells, while
relative control success is already saturated in most cells. We therefore emphasize prediction
under dynamics shifts rather than a large control-success advantage.

\subsubsection{LLM Equation Discovery from Shared Observations}
\label{app:shared-observation-llm}

\lead{Sharing observations acquired by \method enables a same-data comparison
of equation discovery.} We compare \method, LLM-SR, and SR-Scientist on 40
Drawer and 40 Door worlds. The baselines receive \method's final public
observations, but not its proposals, fitted coefficients, or admission
feedback. They issue no additional queries and use their own search and
fitting procedures to generate equations. All three methods use
\texttt{gpt-5.6-sol} with medium reasoning and a cap of 12 model calls per world
over the entire discovery process. Recovery requires mechanism-force
prediction NMSE at most 0.03 on independent motion trajectories, with all
worlds retained in the denominator.

\begin{table}[H]
\caption{\lead{\method achieves higher force-prediction recovery on Door with
shared observations.}}
\label{tab:shared-observation-llm}
\centering
\small
\begin{tabularx}{0.93\linewidth}{ll*{3}{>{\raggedleft\arraybackslash}X}}
\toprule
Task & Method & \shortstack{Force-prediction\\recovery $\uparrow$}
& \shortstack{Median\\NMSE $\downarrow$}
& \shortstack{Mean model\\calls} \\
\midrule
Drawer & \method & 38/40 & 1.871e-07 & 4.100 \\
Drawer & LLM-SR & 37/40 & 2.232e-06 & 12.000 \\
Drawer & SR-Scientist & 39/40 & 5.076e-07 & 3.125 \\
\midrule
Door & \method & 40/40 & 3.995e-08 & 4.700 \\
Door & LLM-SR & 29/40 & 1.067e-04 & 12.000 \\
Door & SR-Scientist & 26/40 & 1.301e-04 & 3.075 \\
\bottomrule
\end{tabularx}
\end{table}

\lead{With shared observations, \method achieves higher force-prediction
recovery on Door.} \method, LLM-SR, and SR-Scientist recover 38, 37, and 39
Drawer worlds, respectively, and 40, 29, and 26 Door worlds. Overall, their
recovery counts are 78/80, 66/80, and 65/80, respectively.

\subsection{Verification Rejects Models Unsupported by Available Evidence}
\label{app:verification}

\subsubsection{Missing Structure and Hidden State}
\label{app:boundary-protocol}

\lead{The equation-and-residual benchmark varies model completeness while matching
observed inputs, noise, sample counts, and domains.} Every regime shares observed distance, speed, action, mass, noise standard
deviation 0.03, sample counts, and ID/shift domains. The known six-term library
contains a constant, distance offset, speed, action divided by mass, a smooth
contact-gated action-over-mass term, and inverse-square distance. Missing terms
are never given to the learner. For incomplete-grammar tests, the residual and gate are trained only after
equation fitting, and boundary labels are evaluated only by the evaluator.

\lead{Model admission and residual localization use separate frozen
criteria.} The equation coefficients are fitted by ridge regression. A two-layer SiLU MLP predicts either the
full target or the residual. A second MLP classifies pseudo-boundary labels
formed when the absolute training residual exceeds four noise standard
deviations. Admission uses held-out ID validation data, requiring equation-only
NMSE at most 0.02 and mean gate probability below 0.10; shifted data serve only
for evaluation.

\begin{figure}[!t]
\centering
\includegraphics[width=\textwidth]{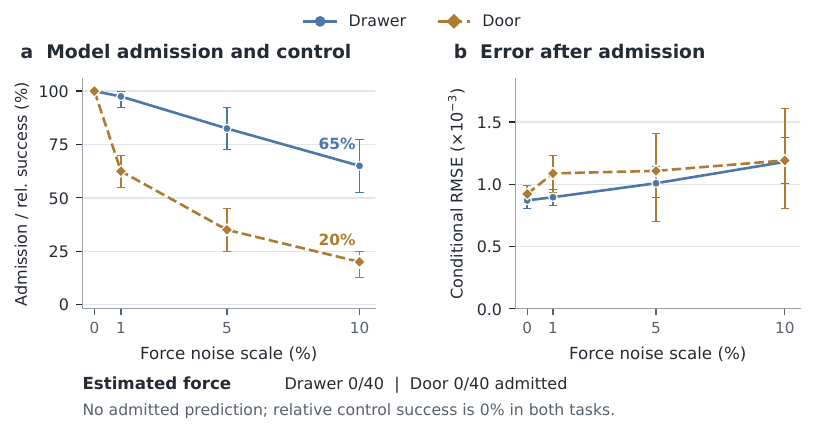}
\caption{\lead{Observation noise reduces admission while retained predictions
remain accurate.} Each task contains 40 worlds across ten seeds. Points show
observed means; whiskers are 95\% seed-bootstrap intervals (10,000 resamples).
Zero noise denotes the force reference. Control uses the original finite-action
relative-success criterion (Table~\ref{tab:robocasa-discrete-control}), not the
inverse-control criterion in Table~\ref{tab:rq4}B. Relative success equals
admission in these runs; rejected worlds have no prediction and receive zero
control success. RMSE is conditional on admission. Force estimation from
motion and known actions (labeled ``Estimated force'') is evaluated separately,
not as another noise level.}
\label{fig:robocasa-observation}
\end{figure}

\begin{table}[!htbp]
\caption{\lead{Residual correction helps missing-term regimes but not
hidden-state failures.} RQ4 reports shifted prediction NMSE; lowest error is
bold.}
\centering
\small
\begin{tabular}{lrrrr}
\toprule
Regime & Equation only & MLP & Always residual & Gated residual \\
\midrule
Complete & \textbf{$4.20{\times}10^{-5}$} & 0.0109 & 0.000345 & \textbf{$4.20{\times}10^{-5}$} \\
Missing smooth & 0.217 & 0.0133 & \textbf{0.00380} & 0.0145 \\
Missing local & 0.0485 & 0.0308 & 0.0191 & \textbf{0.0185} \\
Hidden local & \textbf{0.108} & 0.134 & 0.183 & 0.168 \\
\bottomrule
\end{tabular}
\end{table}

\begin{samepage}
\subsubsection{Observation-Channel Stress Tests}
\label{app:robocasa-observation}

\lead{The observation-channel stress test refits the validated equation structures after
degrading their target signal.} This stress test uses the RoboCasa worlds and control protocol specified in
Appendix~\ref{app:robocasa-protocol}. Noise conditions add 1\%, 5\%, or 10\% scale
noise plus a fixed per-rollout bias. The estimated-force condition replaces
direct force observations with estimates from motion and known control actions.
It estimates nominal acceleration with a 300-tree extra-trees regressor over
position, velocity, and action, then applies a task-specific known-force calibration.
Figure~\ref{fig:robocasa-observation} reports every task and channel. The test is
intended to expose an observability boundary, not to guarantee successful
validation with estimated forces.
\par
\end{samepage}

\lead{The main observation-stress results use the same numerical inverse
solver and absolute success criterion as RQ3.} Table~\ref{tab:rq4}B retains
each channel's fitted equations, admission decisions, and fixed-probe
prediction RMSE. Control success is evaluated over all targets, counting
refusals and unsuccessful executions as failures. It is distinct from the
relative finite-action success shown in Figure~\ref{fig:robocasa-observation}.

\FloatBarrier

\begin{figure}[!t]
  \centering
  \includegraphics[width=\textwidth]{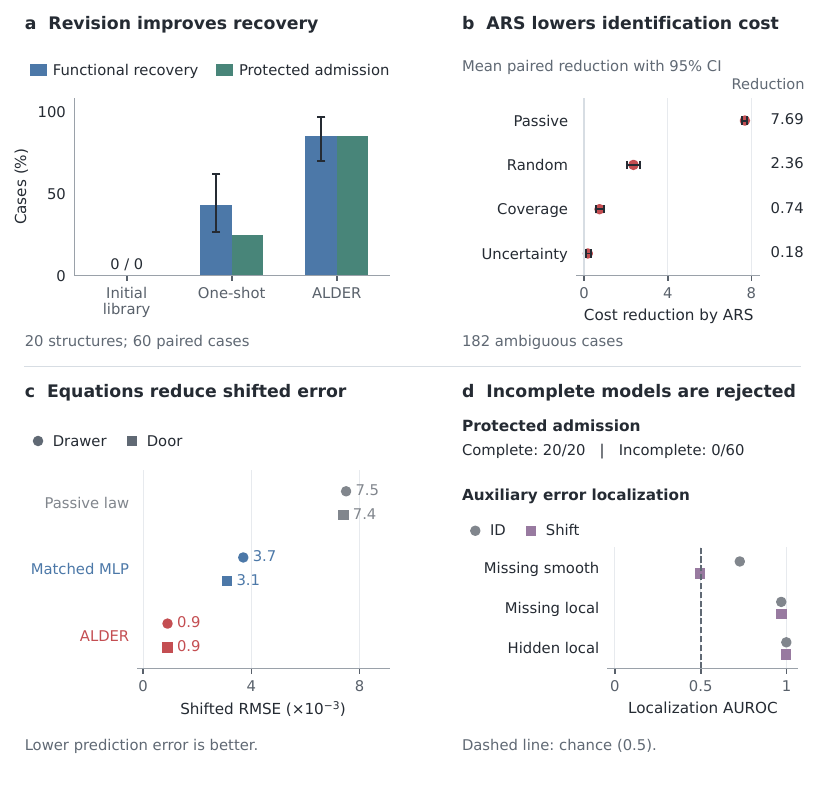}
  \caption{\lead{Closed-loop discovery improves recovery, identification, and
  shifted prediction.} (a) Iterative revision improves
  NovelLaw recovery over the first proposal. (b) Prediction disagreement reduces
  retrospective identification cost with fixed candidate models. (c) Admitted models
  reduce shifted prediction error on two RoboCasa articulations. (d) Verification
  rejects all tested incomplete models; the auxiliary localization diagnostic
  remains near chance for the shifted missing-smooth regime. Error bars show
  95\% bootstrap intervals for functional recovery in (a) and paired identification-cost reductions
  in (b).}
  \label{fig:results}
\end{figure}

\FloatBarrier

\subsection{Equation Discovery Extends Across Systems and Proposal Backbones}
\label{app:extensions}

\subsubsection{External ODE Benchmarks}
\label{app:external-ode}

\lead{The external ODE evaluation covers 122 public dynamical systems.}
We compare 14 method configurations on all 63 ODEBench
systems~\citep{dascoli2024odeformer} and 59 ODEBase
systems~\citep{lueders2022odebase}. The experiments use exact state derivatives. Active methods
start from the same 20-point trajectory and may acquire at most two additional
20-point trajectories. Passive baselines receive \method's final public
fitting and development observations. Failed feasibility attempts are recorded
separately from acquired trajectories.

\lead{The LLM-based methods share a backbone while retaining their own search
procedures and budgets.} All LLM components use \texttt{gpt-5.6-sol} with
medium reasoning and retain method-specific prompts and expression spaces.
SINDy is evaluated with both
restricted and expanded libraries; Table~\ref{tab:rq1}B reports the expanded
version. QBC and BO are reconstructed from published method descriptions.

\lead{A common trajectory evaluator measures prediction and functional
recovery across three settings.} We evaluate reconstruction, generalization
to an unseen initial condition, and continuous future-time extrapolation
without restarting from a withheld future state. NMSE is the sum of squared
prediction errors divided by the target trajectory's squared norm plus
$10^{-10}$. Integration failures receive an NMSE of $10^{30}$ and remain in the
reported medians. Functional recovery requires NMSE at most $10^{-6}$ in all
three settings.

\lead{Symbolic accuracy requires every fitted equation in a system to match
the reference.} A common, method-blind, model-assisted evaluator
compares the fitted equations with the reference equations. Numerical
coefficients remain fixed. Symbolic accuracy is the percentage of matching systems,
rather than an average of partial matches across coordinates. This assessment
is not a formal mathematical proof.

\begin{table}[!htbp]
\caption{\lead{\method recovers 58 of 63 ODEBench systems.} The three NMSE values are medians
over all 63 systems. Functional recovery and symbolic accuracy are evaluated
separately; failures remain in the denominators.}
\label{tab:external-odebench}
\centering
\scriptsize
\setlength{\tabcolsep}{3.5pt}
\renewcommand{\arraystretch}{0.96}
\begin{tabularx}{0.93\linewidth}{l*{5}{>{\raggedleft\arraybackslash}X}}
\toprule
\textbf{Method} & \textbf{Recon. NMSE $\downarrow$} &
\textbf{Gen. NMSE $\downarrow$} & \textbf{OOD NMSE $\downarrow$} &
\textbf{\shortstack{Functional\\recovery $\uparrow$}} & \textbf{Sym. Acc (\%) $\uparrow$} \\
\midrule
\multicolumn{6}{l}{\textbf{\textit{Passive Symbolic Discovery}}} \\
SINDy (expanded) & 1.34e-14 & 1.95e-11 & 1.86e-09 & 33/63 & 25.4 \\
SINDy (restricted) & 2.01e-09 & 1.50e-03 & 3.31e-03 & 14/63 & 14.3 \\
Operon & 3.19e-14 & 3.98e-13 & 4.97e-11 & 39/63 & 22.2 \\
PySR & 1.91e-05 & 8.25e-03 & 1.81e-01 & 11/63 & 12.7 \\
E2E & 4.79e-02 & 1.03e-01 & 4.32e+00 & 1/63 & 0.0 \\
ODEFormer & 2.32e-02 & 6.99e-02 & 8.46e-01 & 1/63 & 0.0 \\
\midrule
\multicolumn{6}{l}{\textbf{\textit{LLM-guided Symbolic Discovery}}} \\
LLM-only & 6.60e-15 & 6.72e-15 & 7.36e-15 & 51/63 & 74.6 \\
LLM-ODE & 8.10e-10 & 1.26e-04 & 2.24e-03 & 20/63 & 4.8 \\
LLM-SR & 5.79e-14 & 5.12e-14 & 4.97e-12 & 39/63 & 55.6 \\
SR-Scientist & 4.11e-14 & 1.52e-13 & 8.46e-11 & 45/63 & 54.0 \\
\midrule
\multicolumn{6}{l}{\textbf{\textit{Active Symbolic Discovery}}} \\
QBC & 1.73e-05 & 1.08e-03 & 5.29e-02 & 14/63 & 22.2 \\
BO & 7.49e-06 & 3.04e-04 & 8.12e-02 & 18/63 & 25.4 \\
LLM-ACES & 3.70e-12 & 2.00e-04 & 1.14e-02 & 23/63 & 25.4 \\
\midrule
\rowcolor{alderrow}
\textbf{\method} & \textbf{1.98e-15} & \textbf{8.16e-16} &
\textbf{5.28e-29} & \textbf{58/63} & \textbf{93.7} \\
\bottomrule
\end{tabularx}

\end{table}

\begin{table}[!htbp]
\caption{\lead{\method recovers 48 of 59 ODEBase systems.} The three NMSE values are medians
over all 59 systems. Functional recovery and symbolic accuracy are evaluated
separately; failures remain in the denominators.}
\label{tab:external-odebase}
\centering
\scriptsize
\setlength{\tabcolsep}{3.5pt}
\renewcommand{\arraystretch}{0.96}
\begin{tabularx}{0.93\linewidth}{l*{5}{>{\raggedleft\arraybackslash}X}}
\toprule
\textbf{Method} & \textbf{Recon. NMSE $\downarrow$} &
\textbf{Gen. NMSE $\downarrow$} & \textbf{OOD NMSE $\downarrow$} &
\textbf{\shortstack{Functional\\recovery $\uparrow$}} & \textbf{Sym. Acc (\%) $\uparrow$} \\
\midrule
\multicolumn{6}{l}{\textbf{\textit{Passive Symbolic Discovery}}} \\
SINDy (expanded) & 1.32e-13 & 8.12e-03 & 2.06e-01 & 8/59 & 5.1 \\
SINDy (restricted) & 1.30e-09 & 3.62e-02 & 1.28e-02 & 3/59 & 3.4 \\
Operon & 1.63e-14 & 2.15e-13 & 1.82e-12 & 35/59 & 23.7 \\
PySR & 1.91e-05 & 2.87e-02 & 9.66e-02 & 4/59 & 5.1 \\
E2E & 2.57e-01 & 1.16e+00 & 2.28e+02 & 0/59 & 0.0 \\
ODEFormer & 2.61e-02 & 4.66e-02 & 9.97e-01 & 0/59 & 0.0 \\
\midrule
\multicolumn{6}{l}{\textbf{\textit{LLM-guided Symbolic Discovery}}} \\
LLM-only & 1.35e-13 & 3.12e-12 & 5.37e-12 & 34/59 & 40.7 \\
LLM-ODE & 3.96e-09 & 1.32e-03 & 2.12e-03 & 12/59 & 1.7 \\
LLM-SR & 6.02e-11 & 1.16e-06 & 1.12e-06 & 26/59 & 32.2 \\
SR-Scientist & 1.23e-12 & 1.36e-11 & 1.26e-10 & 38/59 & 25.4 \\
\midrule
\multicolumn{6}{l}{\textbf{\textit{Active Symbolic Discovery}}} \\
QBC & 7.27e-05 & 8.57e-03 & 8.42e-02 & 7/59 & 8.5 \\
BO & 1.83e-04 & 4.44e-03 & 2.68e-02 & 10/59 & 11.9 \\
LLM-ACES & 2.33e-12 & 2.02e-04 & 1.01e-06 & 19/59 & 23.7 \\
\midrule
\rowcolor{alderrow}
\textbf{\method} & \textbf{2.32e-15} & \textbf{2.31e-16} &
\textbf{9.46e-30} & \textbf{48/59} & \textbf{79.7} \\
\bottomrule
\end{tabularx}

\end{table}

\lead{\method achieves functional recovery and symbolic matches on 106 of the
122 systems.} On ODEBench, the respective counts are 58/63 and 59/63; on
ODEBase, they are 48/59 and 47/59. Both metrics reach 86.9\% overall.
These results show that the harness
can execute across broad ODE collections.

\lead{Fixed-data observation ablations test sensitivity to derivative
estimation and state noise.} We prespecify 12 ODE systems stratified by dataset
and system dimension and compare a fixed-data LLM structure proposer, SINDy,
and PySR. All three methods use the same public trajectories, data splits, and
final trajectory-evaluation procedure.

\lead{Four observation conditions examine exact derivatives, numerical
differentiation, and noise.} We compare exact derivatives, numerical
differentiation of noiseless states, and numerical differentiation after adding
0.1\% or 1\% state noise. The noise is independent Gaussian noise with standard
deviation equal to 0.1\% or 1\% of each state coordinate's standard deviation.
Numerical differentiation consistently uses \texttt{numpy.gradient}, without
additional smoother selection. Functional recovery requires trajectory NMSE at
most $10^{-6}$ in reconstruction, generalization to an unseen initial
condition, and continuous future-time extrapolation; failures remain in the
denominator.

\begin{table}[H]
\caption{\lead{Numerical differentiation and state noise reduce functional
recovery on the tested systems.}}
\label{tab:ode-observation-sensitivity}
\centering
\small
\begin{tabularx}{0.93\linewidth}{>{\raggedright\arraybackslash}Xrrr}
\toprule
Observation condition & \shortstack{LLM structure proposer\\(fixed data)}
& SINDy & PySR \\
\midrule
Exact derivatives & 6/12 & 1/12 & 0/12 \\
Noiseless numerical differentiation & 3/12 & 0/12 & 0/12 \\
Numerical differentiation, 0.1\% state noise & 1/12 & 0/12 & 0/12 \\
Numerical differentiation, 1\% state noise & 0/12 & 0/12 & 0/12 \\
\bottomrule
\end{tabularx}
\end{table}

\lead{Fixed-data structure discovery is sensitive to derivative observation
quality.} The LLM structure proposer's recovery falls from 6/12 with exact
derivatives to 3/12 with noiseless numerical differentiation and declines
further as state noise increases.

\subsubsection{Articulated Geometry and Contact Constraints}
\label{app:robot-discovery}

\lead{Door and Peg test articulated geometry and non-smooth constraints
without requiring multi-round revision.} Every formal Agent run succeeds on
its first proposal, so these tasks serve as breadth checks rather than
evidence that revision is needed. Adroit Door uses
episode-disjoint angle ranges to test additive approximations fitted at low angles. We use
the standard Minari \texttt{D4RL/door/human-v2}
conversion~\citep{rajeswaran2018dapg,fu2020d4rl}, containing 25
human demonstrations and 6,729 transitions. For each of five episode
partitions, public fit and development rows contain hinge angles no larger than
$0.65$ rad. Protected ID holds out episodes in the same range; protected OOD
holds out episodes at hinge angles of at least $0.90$ rad. Inputs are current
and initial hinge/latch angles, and the target is handle $x$ displacement.
Admission requires development, ID, and OOD NMSE below 0.04, 0.05, and 0.02,
respectively, plus at least 90\% improvement over the zero-displacement
baseline on both protected splits.

\lead{Every admitted Door equation contains the composed geometric dependence
needed for high-angle extrapolation.} The agent cannot see protected rows or the harness-held composed
trigonometric basis. Every first proposal passes admission. The admitted
equations use hinge sine/cosine displacement together with a latch term rotated
by the hinge, rather than adding the two joints independently. Table
~\ref{tab:door-full} reports all registered baselines.

\begin{table*}[!ht]
\caption{\lead{One-shot joint equations extrapolate to unseen Door angles.}
Adroit Door uses five episode partitions. Errors are handle-$x$ MAE in cm;
lower is better. OOD contains only unseen high-angle states.}
\label{tab:door-full}
\centering
\small
\begin{tabular}{lrrrr}
\toprule
Method & Public dev. & Protected ID & High-angle OOD & Admitted \\
\midrule
Zero displacement & 6.671 & 6.501 & 28.866 & 0\% \\
Linear joints & 0.810 & 0.813 & 11.660 & 0\% \\
Polynomial SINDy-3 & 0.025 & 0.027 & 1.455 & 40\% \\
MLP & 0.047 & 0.051 & 10.546 & 0\% \\
Additive trigonometric library & 0.068 & 0.065 & 2.626 & 0\% \\
Full composed reference & 0.020 & 0.021 & 0.568 & 100\% \\
One-shot agent / \method & 0.046 & 0.047 & \textbf{0.339} & 100\% \\
\bottomrule
\end{tabular}
\end{table*}

\lead{Peg interventions test contact constraints beyond correlations with task
progress.} The 1,000 official motion-planning demonstrations contain 149,055 labeled
states, but follow a narrow approach path. A linear classifier obtains about
0.996 balanced accuracy on those states by relying on task progress. We
therefore execute 56 orthogonal state interventions for every official episode
seed. Seven balanced probe families isolate success, axial failure,
positive/negative $y$ failure, positive/negative $z$ failure, and simultaneous
failures; half of the relevant probes lie near a boundary. ManiSkill executes
all 56,000 probes and supplies the only public target, a binary success label.

\lead{Peg evaluation separates central-radius fitting from protected episode and
radius shifts.} Public fit uses 40 central-radius episodes (2,240 labels), public development
uses 20 (1,120), protected ID uses 40 (2,240), and radius OOD uses 80 (4,480)
split equally between disjoint small- and large-hole ranges. Admission requires
at least 0.98 balanced accuracy and 0.95 success F1 on development, protected
ID, and radius OOD. Across five partitions, every first proposal produces an
equivalent three-branch constraint and passes admission (Table
~\ref{tab:peg-full}). The continuous simulator margin is never released to the
agent or learned baselines.

\begin{table*}[!ht]
\caption{\lead{One-shot constraints generalize across held-out peg-hole
radii.} PegInsertionSide provides binary simulator labels. All metrics except
``Admitted'' are balanced accuracy or success F1; higher is better.}
\label{tab:peg-full}
\centering
\scriptsize
\setlength{\tabcolsep}{3.5pt}
\begin{tabular}{lrrrrr}
\toprule
Method & Public dev. & Protected ID & Radius OOD & OOD success F1 & Admitted \\
\midrule
Always failure & 0.500 & 0.500 & 0.500 & 0.000 & 0\% \\
Linear & 0.557 & 0.553 & 0.552 & 0.266 & 0\% \\
Polynomial logistic-3 & 0.769 & 0.767 & 0.754 & 0.450 & 0\% \\
MLP & 0.905 & 0.901 & 0.796 & 0.605 & 0\% \\
Additive absolute rule & 0.649 & 0.646 & 0.638 & 0.369 & 0\% \\
Full non-smooth reference & 1.000 & 1.000 & 1.000 & 1.000 & 100\% \\
One-shot agent / \method & \textbf{1.000} & \textbf{1.000} & \textbf{1.000} & \textbf{1.000} & 100\% \\
\bottomrule
\end{tabular}
\end{table*}

\begin{samepage}
\subsubsection{Proposal-Backbone Robustness}
\label{app:backbones}

\lead{The open-weight ablation changes only the proposal backbone while
fixing the experimental framework.} All models use the same stratified subset
of 12 RoboCasa worlds across three paired seeds, with identical public
evidence, selector, coefficient fitter, admission criteria, and query budget.
Table~\ref{tab:robocasa-open-models} compares the primary GPT-5.6 backend with
seven valid open-weight backbones on this subset. Failed proposals remain in
the denominators of admission and term-support recovery rates.
\par
\end{samepage}

\begin{samepage}
\lead{ALDER discovers validated equations with multiple open-weight proposal models.}
Across the seven open-weight backbones, protected admission ranges from 58.3\%
to 100.0\% and strict term-support recovery from 33.3\% to 75.0\%. The primary
GPT-5.6 backend obtains 100.0\% admission and 83.3\% term-support recovery on the
identical subset. All backbones record zero safety violations. The pattern is
not monotonic in total or active parameter count: the dense 27B model has the
highest open-weight term-support recovery, whereas the sparse 80B model has the
lowest admission.
\par
\end{samepage}

\begin{table*}[!ht]
\caption{\lead{\method discovers validated equations with multiple open-weight
backbones.} On 12 matched RoboCasa worlds, the first row is the primary backend;
the remaining seven are open-weight ablations. Queries are medians with
non-admitted runs assigned budget plus one; NMSE is the geometric mean over
finite predictions.
Total/active parameter counts follow public model declarations, while the
primary parameter count is not public. Tokens are API-reported mean request
totals and are not a quality-normalized cost metric. ``Support'' uses the
same strict optional-term recovery criterion as the primary study.}
\label{tab:robocasa-open-models}
\centering
\scriptsize
\setlength{\tabcolsep}{2.8pt}
\begin{tabular}{lrrrrrrr}
\toprule
Backbone & Total/active B & Admitted & Support & Queries & Pred. cover & NMSE & Tokens \\
\midrule
\textbf{GPT-5.6 (primary)} & not public & 100.0\% & \textbf{83.3\%} & 0.5 & 100.0\% & $7.85\times10^{-8}$ & 345788 \\
Qwen3.5-9B & 9/9 & 100.0\% & 66.7\% & 1.0 & 100.0\% & $3.34\times10^{-7}$ & 35082 \\
GPT-OSS-20B & 21/3.6 & 100.0\% & 41.7\% & 1.0 & 100.0\% & $3.16\times10^{-6}$ & 43037 \\
Qwen3.8-27B & 27/27 & 91.7\% & 75.0\% & 0.5 & 100.0\% & $8.77\times10^{-7}$ & 56292 \\
Qwen3.6-35B-A3B & 35/3 & 91.7\% & 41.7\% & 1.0 & 100.0\% & $3.93\times10^{-6}$ & 54832 \\
Qwen3-Coder-Next & 80/3 & 58.3\% & 33.3\% & 1.5 & 100.0\% & $5.85\times10^{-5}$ & 78094 \\
GPT-OSS-120B & 117/5.1 & 100.0\% & 41.7\% & 1.0 & 100.0\% & $2.67\times10^{-6}$ & 31470 \\
Qwen3-235B-A22B & 235/22 & 100.0\% & 50.0\% & 0.5 & 100.0\% & $3.04\times10^{-6}$ & 78166 \\
\bottomrule
\end{tabular}

\end{table*}

\FloatBarrier

\subsection{Statistical Protocols and Model Settings Support Reproducibility}
\label{app:audit}

\subsubsection{Statistical Protocols}

\lead{Statistical comparisons preserve the matched unit appropriate to each
experiment.} All final comparisons are paired by generated case or robot task. Interaction
counts use percentile-bootstrap intervals. RQ1 recovery-rate intervals cluster
the bootstrap over 20 structure recipes, retaining all three parameter/data
instances of a sampled recipe; its log-NMSE comparisons are paired by case.
Wilcoxon signed-rank and exact McNemar tests are reported as secondary paired
diagnostics.

\subsubsection{Proposal Models and Resource Use}

\lead{NovelLaw's primary proposal backend was identified as GPT-5.6 at
evaluation time.} The primary runs used the default backend, which resolved
to \texttt{gpt-5.6-sol} with low reasoning effort; auxiliary revision and
random-query runs explicitly specified \texttt{gpt-5.6-sol}.

\lead{We summarize resource use through token counts and cumulative agent
time.} Table~\ref{tab:rq1-cost} reports usage over 60 NovelLaw cases per
variant; on RoboCasa, mean API-reported usage is 413,528 tokens per Drawer
world and 323,520 per Door world.

\begin{table}[h]
\caption{\lead{Each full agent variant uses one invocation per case.} RQ1
usage is summed over 60 cases. One-shot is derived from \method's stored first
round and incurs no additional call.}
\label{tab:rq1-cost}
\centering
\small
\begin{tabular}{lrrrrr}
\toprule
Method & Calls & Proposals & Queries & Input tokens & Agent h \\
\midrule
Revision-only & 60 & 138 & 0 & 31.8M & 1.97 \\
Random-query & 60 & 117 & 1,368 & 30.9M & 2.15 \\
\method & 60 & 125 & 1,560 & 34.1M & 2.22 \\
\bottomrule
\end{tabular}
\end{table}

\FloatBarrier

\end{document}